\pdfoutput=1

\documentclass[11pt]{article}

\usepackage{acl}

\usepackage{times}
\usepackage{latexsym}

\usepackage[T1]{fontenc}

\usepackage[utf8]{inputenc}

\usepackage{microtype}

\usepackage{times}
\usepackage{latexsym}
\usepackage{epsfig}
\usepackage{graphicx}
\usepackage{amsmath}
\usepackage{amssymb}
\usepackage{xspace}
\usepackage{amstext}
\usepackage{multirow}
\usepackage{tabularx} 
\usepackage{booktabs}   
\usepackage{subcaption}
\usepackage{colortbl}
\usepackage{hyperref}
\usepackage{todonotes}
\usepackage{placeins}
\usepackage{wrapfig}
\usepackage{pifont}
\usepackage{siunitx}
\usepackage{soul}

\newcommand{\rparagraph}[1]{\vspace{1.4mm}\noindent\textbf{#1.}}

\title{Does Object Grounding Really Reduce Hallucination of Large Vision-Language Models?}
\author{\textbf{Gregor Geigle$^{12}$ \quad \quad Radu Timofte$^{2}$ \quad \quad Goran Glava\v{s}$^{1}$} \\
  $^{1}$W{\"u}NLP, $^{2}$Computer Vision Lab, CAIDAS, University of W{\"u}rzburg,  \\
  \texttt{gregor.geigle@uni-wuerburg.de}}

\begin{document}
\maketitle
\begin{abstract}
Large vision-language models (LVLMs) have recently dramatically pushed the state of the art in image captioning and many image understanding tasks (e.g., visual question answering). LVLMs, however, often \textit{hallucinate} and produce captions that mention concepts that cannot be found in the image. These hallucinations erode the trustworthiness of LVLMs and are arguably among the main obstacles to their ubiquitous adoption. 
Recent work suggests that addition of grounding objectives---those that explicitly align image regions or objects to text spans---reduces the amount of LVLM hallucination. Although intuitive, this claim is not empirically justified as the reduction effects have been established, we argue, with flawed evaluation protocols that 
(i) rely on data (i.e., MSCOCO) that has been extensively used in LVLM training and (ii) measure hallucination via question answering rather than open-ended caption generation.
In this work, in contrast, we offer the first systematic analysis of the effect of fine-grained object grounding on LVLM hallucination under an evaluation protocol that more realistically captures LVLM hallucination in open generation. 
Our extensive experiments over three backbone LLMs reveal that grounding objectives have little to no effect on object hallucination in open caption generation.
\end{abstract}

\section{Introduction}
Large Vision-Language Models (LVLMs)
 have recently displayed impressive image understanding abilities \cite[\textit{inter alia}]{li_blip-2_2023,liu_visual_2023,bai_qwen-vl_2023,fini_improved_2023,openai_gpt-4_2023,anil_gemini_2023}.
Their widespread adoption, however, is hindered
by \textit{object hallucination}
in which the LVLMs---similar to ``general'' hallucination of LLMs \cite{zhang_sirens_2023}---``invent'' objects (or attributes of or relations between objects) not present in the image.

A range of methods have recently been proposed to address LVLM hallucination such as
modified decoding strategies \cite{leng_mitigating_2023,huang_opera_2023}, post-hoc removal of hallucinated content \cite{yin_woodpecker_2023,zhou_analyzing_2023}, 
or reinforcement learning \cite{sun_aligning_2023,zhao_beyond_2023,gunjal_detecting_2023,yu_rlhf-v_2023}. Most of these approaches, however, either increase inference cost
or need expensive additional training and/or data, impeding their ubiquitous applicability.

A recent line of work \cite{chen_shikra_2023,you_ferret_2023,pramanick_jack_2023} has suggested that including 
\textit{grounding objectives}---e.g., based on referring expressions \cite{kazemzadeh_referitgame_2014} where textual descriptions of image regions have to be grounded to the respective parts of the image---into the LVLM training reduces object hallucination. The claim is intuitive: 
region-level objectives demand finer-grained image understanding than the `global' image captioning (\textit{de facto} the main training objective of LVLMs), as demonstrated in visiolinguistic compositionality \cite{bugliarello_measuring_2023-1}. Such objectives should thus, intuitively, discourage models from generating content they cannot ground in the image.
%
%
Intuition aside, the empirical support for the claim that grounding objectives reduce LVLM hallucination is weak and mainly limited to question-answering (QA) style of evaluation in which the model is explicitly asked about existence of objects in an image \cite{li_evaluating_2023};
we argue that this evaluation protocol poorly aligns with real-world \textit{free-form} text generation tasks---primarily open image captioning---for which there is no empirical evidence yet that object grounding reduces hallucination. 

%

\rparagraph{Contributions}
In this work, we perform the first comprehensive analysis of the effects that grounding objectives have on LVLM object hallucination in open (i.e., free-form) image captioning, addressing the shortcomings of existing hallucination evaluation protocols. Concretely, we measure the effect of adding two popular grounding objectives as additional objectives to standard image captioning-based training of LVLMs: (1) the \textit{referring expressions (RE)} objective asks the model to generate the bounding box of the region that corresponds to a textual description and vice versa; 
whereas (2) the \textit{grounded captioning (GC)} objective demands that the model generates image descriptions with interleaved (relative coordinates of) bounding boxes for mentioned objects.
We then compare the extent of hallucination for LVLM variants trained with and without these grounding objectives. To this end, we compare the hallucination measures based on question answering (QA) \cite{li_evaluating_2023} against free-form metrics for open captioning \cite{rohrbach_object_2018,jing_faithscore_2023}.
Critically, observing that (1) existing evaluation measures and protocols \cite{rohrbach_object_2018,li_evaluating_2023} rely on MSCOCO \cite{lin_microsoft_2014} and (2) MSCOCO data is part of the training mix for most LVLMs, we argue that existing measures are likely to underestimate LVLM hallucinate; we thus extend our hallucination evaluation protocol to out-of-distribution data that LVLMs will not have seen in training.


\rparagraph{Findings} 
Our experiments with three different LLM backbones show that, under a sound evaluation protocol, including grounding objectives---referring expressions and grounded captioning---to LVLM training has little to no effect on object hallucination, both in QA-based evaluation and open-ended captioning.
Enforcing generation of \textit{grounded captions} at inference time, on the other hand, slightly reduces object hallucinations but the effect is small and comes at the cost of (slight) reduction in caption detailedness. A qualitative inspection of grounded captions also confirms that forcing model to generate a bounding box for mentioned objects most often does not prevent it from hallucinating content. In sum, we find that grounding objectives fail to meaningfully reduce LVLM hallucination, calling for novel methodological proposals towards hallucination reduction.


\begin{figure*}[t]
    \centering
    \includegraphics[width=0.9\linewidth]{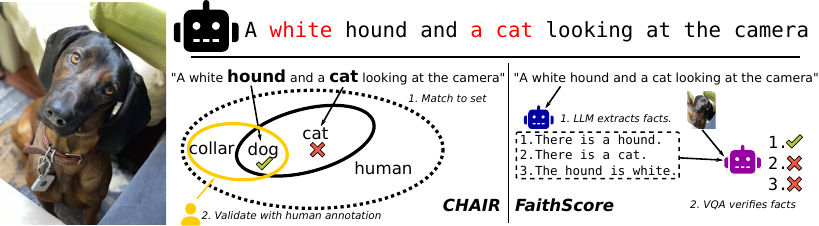}
    \caption{
    \textbf{CHAIR} and \textbf{FaithScore} are used to measure hallucinations in open caption generation with LVLMs. \textbf{CHAIR} relies on human object annotation (over a fixed set) to identify objects and check if they are hallucinated.
    \textbf{FaithScore} first uses an LLM to convert captions into facts which are then verified by a VQA model.
    }
    \label{fig:metrics}
    \vspace{-1em}
\end{figure*}

\section{Grounding Objectives in LVLMs}
\label{sec:grounding}
Grounding objectives seek to align natural language expressions 
with regions in the image. 
These objectives either take image regions as input, in the form of a bounding box and predict corresponding language expressions or produce such regions as output. Many recent LVLMs have been trained with grounding tasks in their training mix alongside standard tasks like captioning and VQA \cite{liu_improved_2023,bai_qwen-vl_2023,wang_cogvlm_2023}; other models have been designed specifically for expression grounding and trained with grounding objectives only \cite{chen_shikra_2023,you_ferret_2023,pramanick_jack_2023,zhang_gpt4roi_2023,peng_kosmos-2_2023,chen_position-enhanced_2023,zhao_bubogpt_2023}. 

\paragraph{Objectives.}
Our investigation focuses on the two arguably most popular grounding objectives, commonly part of LVLM training: referring expressions \cite{kazemzadeh_referitgame_2014} and grounded captioning \cite{plummer_flickr30k_2015}. 

\textit{Referring expressions} is the standard grounding objective, included in training of nearly all LVLMs. Given a natural language description (of a region), the model has to ground it to the correct image region. As is common practice, we also use the inverse task, that is, generation of the natural language description for the given image region. 

\textit{Grounded captioning} is the task of generating an image caption in which the locations of regions 
for mentioned objects are interleaved in the caption (see Figure~\ref{figure:example_grounded} for examples).
In theory, such explicit grounding is expected to result in closer adherence to the image content and reduce hallucinations. 

Other grounding objectives have been proposed for LVLMs training, such as question answering with image regions in the input or output \cite{zhu_visual7w_2016}; 
these, however, are outside the scope of our study, because we focus on the effects of grounding on hallucination primarily in free-form captioning. 

\rparagraph{Encoding regions}
Different approaches exist for representing image regions for the LVLMs. Most commonly, regions are represented as bounding boxes using either (relative) coordinates in ``plain text'' \cite{liu_improved_2023,chen_shikra_2023,bai_qwen-vl_2023,wang_cogvlm_2023} (e.g., ``$[0.10, 0.05, 0.64, 1.00]$''; the coordinates are treated as text and tokenized with the tokenizer of the corresponding LLM) or with learned embeddings that correspond to a fixed-size rasterization of the image \cite{peng_kosmos-2_2023,you_ferret_2023,pramanick_jack_2023}.
In this work, we adopt the former region representation, i.e., relative coordinates as text, as this avoids introducing additional trainable parameters to the model.  


\section{Measuring Object Hallucination}
\label{sec:metrics}
LVLM object hallucination is evaluated via two main protocols: (1) in QA-based evaluation, where models answer questions about object existence in the image \cite{li_evaluating_2023} and (2) in open generation, usually image captioning  \cite{rohrbach_object_2018,wang_evaluation_2023,jing_faithscore_2023}. The latter is arguably more indicative of models' tendency to hallucinate ``in the wild'' (i.e., in various real-world applications)
 but it is also a more difficult setup for automatic evaluation.
In contrast, QA-based evaluation is straightforward, 
but an untested proxy for actual hallucination in generative tasks.



\rparagraph{QA-Based Hallucination Evaluation}
POPE \cite{li_evaluating_2023} is the \textit{de facto} standard benchmark for QA-based hallucination evaluation.
Relying on images annotated with objects from MSCOCO \cite{lin_microsoft_2014}, the benchmark consists of \textit{yes}/\textit{no} questions about object existence (``\textit{Is there X in the image?}''). The negative questions---about objects \textit{not} in the image---are generated in three different ways using: i) objects randomly selected from the total pool of objects that exist in the dataset (\textit{random}); ii) the most frequently annotated objects in the dataset (\textit{popular}); iii) objects with high co-occurrence to the image's actual objects (\textit{adversarial}), as co-occurrence statistics are a common cause of hallucinations \cite{rohrbach_object_2018,biten_let_2022,li_evaluating_2023,zhou_analyzing_2023}. 
The performance metric is accuracy, i.e., the percentage of correctly answered questions. 

\rparagraph{Open Hallucination Evaluation}
We focus on two popular meatrics for quantifying hallucination in open caption generation: CHAIR \cite{rohrbach_object_2018} and FaithScore \cite{jing_faithscore_2023}, illustrated in Figure~\ref{fig:metrics}). The two metrics identify hallucination in different ways: by complementing them with one another, 
we mitigate the risk of our findings merely being an artifact of a single (imperfect) evaluation metric.
Both metrics can also indirectly quantify how \textit{informative} and descriptive the generated captions are.  
As our result will show (\S\ref{ssec:results}), there exists a tradeoff between faithfulness/hallucination and informativeness of the captions. 
We thus argue that the hallucination metrics should be contextualized with the measures of informativeness: factually correct but uninformative captions are as undesired as captions with hallucinated information.   

\noindent\textbf{CHAIR} detects hallucinated objects using the set of 80 object classes from MSCOCO \cite{lin_microsoft_2014} with which the images are annotated. 
Words from the captions are matched---using exact string matching---against the class names, augmented with synonyms. The resulting list of matched objects is then cross-referenced against the gold list of annotated objects and all matched but not annotated objects are considered hallucinations. Two scores are produced over the dataset: (1) CHAIR$_i$ divides the total number of hallucinated objects across all captions with the total number of detected objects; (2) CHAIR$_s$ is the proportion of images in the dataset for which the caption contains at least one object hallucination. CHAIR$_s$ is less than ideal for longer captions as they are more likely to contain at least one hallucination;
such a binary caption-level measure would hide potentially substantial differences in hallucination rates between models. Because of this, we adopt only CHAIR$_i$ in this work.        
Following \citet{zhai_halle-switch_2023}, we additionally report the average number of matched objects 
per caption as well as the gold object coverage (i.e., the average percentage of annotated objects mentioned in the caption) as measures of caption \textit{informativeness}.

CHAIR unfortunately comes with two major shortcomings. First, it is based on MSCOCO images and object annotations which are widely used in a range of derivative datasets leveraged for training LVLMs \cite{goyal_making_2017,kazemzadeh_referitgame_2014,mao_generation_2016,liu_visual_2023}. This makes LVLMs \textit{a priori} less likely to hallucinate on MSCOCO images, which means that CHAIR is likely overly optimistic about (i.e., it underestimates) the amount of LVLM hallucination ``in the wild''. We thus propose to extend CHAIR to an out-of-distribution dataset, one that ideally also comes with a larger set of object classes.
Second, CHAIR relies on exact string matching between caption words and synonym sets of the object classes. Adapting vanilla CHAIR based on string matching to a larger set of object classes would, however, require significant manual effort, as one would have to (1) create a curated list of synonyms for all new classes (without overlap between related classes) to correctly account for recall and (2) inspect examples and create special rules for edge cases to limit false positives (e.g., add `baby X' synonyms to all animal classes `X' in order not to falsely match the 'person' class).
Addressing both issues simultaneously, we propose semantic matching between the caption and object classes as an 
alternative to string matching for large sets of object classes. Our extension, dubbed \textbf{CHAIR-MEN} (from \textbf{CHAIR} with \textbf{M}atching using \textbf{E}mbeddings of \textbf{N}oun phrases) (1) extracts all noun phrases from the generation,\footnote{With spaCy v3 \textsc{en\_core\_web\_sm}} (2) embeds the extracted phrases as well as classes names 
with a pretrained sentence encoder \cite{reimers_sentence-bert_2019}\footnote{\textsc{BAAI/bge-base-en-v1.5} \cite{xiao_c-pack_2023}} and (3) makes matching decisions based on cosine similarity between obtained embeddings: to each noun phrase, we assign (i) the class amongst the image's objects with the most similar embedding, if cosine exceeds a threshold $t_1$, (ii) the class amongst the other objects (i.e., not present in the image) with the most similar embedding, if cosine exceeds a threshold $t_2$, or otherwise (iii) no object class. Matching first only against the image's objects makes false negatives from a semantically related object not in the image less likely.
We calibrate the thresholds ($t_1=0.73, t_2=0.78$) 
by trying to match the scores that vanilla CHAIR produces on MSCOCO, as an established measure for that dataset. 

\noindent\textbf{FaithScore} \cite{jing_faithscore_2023}, a model-based hallucination metric, is designed with finer-grained evaluation in mind: it does not only consider objects/entities but also other aspects that models can hallucinate about (specifically: color, relation, count, and `other' attributes), without the need for human annotation. FaithScore computation is a 2-stage process that relies: (1) on an LLM to extract `atomic facts' from the generated text, phrasing them as statements (e.g., \textit{``There is a man''}) the factuality of which, in the context of the image, is then (2) verified with a VQA model (question: ``Is the following statement correct?''). The final score is then simply the proportion of positive answers given by the VQA model. We additionally report the average number of facts produced by the LLM as a measure of informativeness of generated captions.
The original work of \citet{jing_faithscore_2023} relies on GPT-4 to extract facts but this is too expensive for our evaluation; instead, we use a smaller LLM\footnote{Llama3-8B-Instruct \cite{aimeta_llama_2024}; inference done with vLLM \cite{kwon_efficient_2023} for speed} after verifying that it successfully follows task instructions. We use OFA \cite{wang_ofa_2022} as the VQA model for FaithScore, as it is much faster and only marginally less accurate than Llava-1.5 \cite{liu_improved_2023} according to \citet{jing_faithscore_2023}. 

\rparagraph{Caption Quality Metrics}
Next to the hallucination measures, we add the following two standard metrics to monitor how grounding objectives affect the general caption quality:
\textbf{CIDEr} \cite{vedantam_cider_2015} is a measure based on n-gram overlap with a 
set of reference captions. \textbf{CLIPScore}, a reference-free metric, is the cosine similarity between the image and caption embeddings, produced
by a CLIP model \cite{radford_learning_2021-1}.\footnote{We use \textsc{ViT-B-16-SigLIP-256} \cite{zhai_sigmoid_2023-1}}

\section{Experimental Setup}

We comprehensively analyze the effect of grounding objectives on LVLM hallucination. For the sake of transferability and robustness of our findings, our experimental core, namely the model architecture and training procedure, follows established practices as closely as possible. All model instances are trained according to the same protocol, that is, we control for everything other than the effect of grounding, i.e., inclusion/exclusion of grounding data during training.
We primarily focus on measuring hallucination in open-ended image captioning as this, we argue, better reflects LVLM's hallucination in real-world applications; for completeness and comparison of evaluation protocols, we also perform the QA-based evaluation with POPE. 
We benchmark LVLMs for hallucinations in two different caption generation scenarios: (1) in \textit{standard} image captioning, with expected caption length of 1-2 sentences (as in MSCOCO), and (2) \textit{grounded} image captioning (with standard length), where the LVLM is explicitly prompted to interleave region coordinates into the caption.
In the Appendix~\ref{sec:appendix:long}, we also provide results for \textit{long} (i.e., detailed, descriptive) caption generation.

\rparagraph{Evaluation Datasets}
Despite the previously mentioned shortcomings, \textbf{MSCOCO} \cite{lin_microsoft_2014} remains the primary dataset for evaluating LVLM hallucination in the literature, both with QA-based and free-form generation metrics/protocols \cite{rohrbach_object_2018,li_evaluating_2023}. 
We thus include MSCOCO but complement it with the \textbf{Objects365} (\textbf{O365}) \cite{shao_objects365_2019-1} dataset which comes with a much larger inventory of object classes (365 classes in total, including the 80 MSCOCO classes) and, consequently, more object annotations per image. We evaluate on 5000 and 5386 images from test portion of MSCOCO and validation portion of O365, respectively.\footnote{We have additionally considered Open Images \cite{kuznetsova_open_2020-1}, Visual Genome (VG) \cite{krishna_visual_2017}, and LVIS \cite{gupta_lvis_2019} as datasets with gold object annotations but ultimately decided against their inclusion due to insufficient object coverage in annotations (i.e., not all objects are annotated in every image).} 
%
For the POPE evaluation,
 we generate two new test sets from O365, 
each with 1500 examples (matching MSCOCO POPE):
\texttt{O365/COCO} uses only the 80 classes from MSCOCO, and \texttt{O365/non-COCO} utilizes the remaining 285 classes.


\rparagraph{LVLM Architecture}
We adopt the typical LVLM architecture: (1) images are encoded by an image encoder, (2) projected by an alignment module into the LLM embedding space, and (3) prepended to the embeddings of textual tokens \cite{liu_improved_2023}.
For the alignment module, we adopt as default the projection by \citet{chu_mobilevlm_2024}, which uses a 2-layer MLP followed by a pooling layer.
We also experiment with a resampler \cite{li_blip-2_2023,bai_qwen-vl_2023,alayrac_flamingo_2022}, which
learns to encode the visual information from the image in a set of trainable query embeddings; specifically, we use a 3-layer perceiver-resampler \cite{alayrac_flamingo_2022} with 32 query tokens. 
We leverage the OpenAI CLIP ViT-L/14-224 \cite{radford_learning_2021} as the image encoder.
We experiment with three different LLM backbones: Vicuna 1.5 7B \cite{chiang_vicuna_2023}, Llama-3 8B (instruct) \cite{aimeta_llama_2024}, and Phi-3-mini \cite{abdin_phi-3_2024}.
The LLM parameters are frozen and 4-bit quantized \cite{dettmers_qlora_2023-1}; instead of direct LLM updates, we learn the LoRA adapters \cite{hu_lora_2022-2} for all parameter matrices of the LLM.

\rparagraph{Pre-Training}
We pre-train the alignment module---and only the alignment module (all other parameter frozen)---on image-caption data. For this, we use the 560k examples from \citet{liu_improved_2023}.

\rparagraph{Training Mix}
LVLMs are generally instruction-trained on a mix of tasks and datasets.
The mix we adopt reflects the main goal of our study:
 to isolate the effect of grounding objectives on LVLMs hallucination. 
We thus include the following tasks: 


\noindent \textit{1. Standard image captioning}: we train on the MSCOCO captions (400k examples);

\noindent \textit{2. Long captioning}: we use \textsc{Llava-Detailed} \cite{liu_visual_2023} with 23k long captions generated by GPT-4 on the basis of (short) MSCOCO reference captions and gold object annotations;

\noindent \textit{3. VQA}: we select from VQAv2 \cite{goyal_making_2017} all 170k yes/no questions. VQA is only added to the training mix for the sake of QA-based hallucination evaluation with POPE;\footnote{Without VQA in the training mix, the LVLMs do not follow the POPE task instruction.} 

\noindent \textit{4. Referring expressions} (see \S\ref{sec:grounding}):
we combine 
RefCOCO \cite{kazemzadeh_referitgame_2014,mao_generation_2016} (320k examples) and Visual Genome \cite{krishna_visual_2017} (we sample 320k examples); 

\noindent \textit{5. Grounded captioning} (see \S\ref{sec:grounding}): we use 
Flickr30k-Entities \cite{plummer_flickr30k_2015} (150k examples).

We name our LVLM model variants based on their respective training mix. The \texttt{Base} LVLM has been trained only on non-grounding tasks (1-3); addition of the referring expressions and grounded captioning tasks is indicated with \texttt{+RE} and \texttt{+GC}, respectively. For brevity, we provide further training and inference details in the Appendix \ref{sec:appendix:train}. 
By default, we use the pooled MLP projection from \citet{chu_mobilevlm_2024} for all models.
Additionally, we train a Vicuna-based model with the perceiver-resampler, which we denote with \texttt{(Perc)}.

\begin{table*}[t]
    \centering
    \footnotesize
     \def\arraystretch{0.8}
     \resizebox{0.99\linewidth}{!}{


\begin{tabular}{lrrrrrrrrr}
\toprule
& \multicolumn{3}{c}{\bf MSCOCO} & \multicolumn{3}{c}{\bf O365/COCO} & \multicolumn{3}{c}{\bf O365/non-COCO} \\
\bf Model & \bf rand. & \bf pop. & \bf adv. & \bf rand. & \bf pop. & \bf adv. & \bf rand. & \bf pop. & \bf adv. \\
\cmidrule(lr){1-1} \cmidrule(lr){2-4} \cmidrule(lr){5-7} \cmidrule(lr){8-10}
    \texttt{Llama-3} \texttt{Base} & 86.87 &  81.73 &      75.83 &     \textbf{83.13} &      70.47 &   65.63 &     \textbf{78.53} &      66.13 &   58.20 \\
    \texttt{Llama-3} \texttt{+GC} & 86.83 &  82.43 &      78.90 &     81.87 &      71.60 &   68.50 &     77.57 &      \textbf{67.70} &   60.37 \\
\texttt{Llama-3} \texttt{+RE} & 84.10 &  81.87 &      \textbf{79.93} &     76.07 &      \textbf{73.10} &   \textbf{71.73} &     70.53 &      67.07 &   \textbf{64.57} \\
\texttt{Llama-3} \texttt{+RE+GC} & \textbf{84.70} &  \textbf{83.77} &      \textbf{79.93} &     75.47 &      71.00 &   69.73 &     67.63 &      64.50 &   61.27 \\
\cmidrule(lr){1-1} \cmidrule(lr){2-4} \cmidrule(lr){5-7} \cmidrule(lr){8-10}
      \texttt{Phi-3} \texttt{Base} & 87.17 &  85.30 &      81.87 &     81.57 &      77.57 &   73.73 &     \textbf{79.10} &      \textbf{74.77} &   66.40 \\
      \texttt{Phi-3} \texttt{+GC} & 85.30 &  83.73 &      81.80 &     78.93 &      75.53 &   73.47 &     72.43 &      69.50 &   65.80 \\
  \texttt{Phi-3} \texttt{+RE} & 86.43 &  \textbf{85.50} &      \textbf{83.50} &     78.93 &      76.20 &   \textbf{74.10} &     75.17 &      72.40 &   \textbf{68.83} \\
  \texttt{Phi-3} \texttt{+RE+GC} & \textbf{87.57} &  85.43 &      81.77 &     \textbf{84.63} &      \textbf{78.27} &   74.00 &     77.03 &      74.30 &   68.30 \\
\cmidrule(lr){1-1} \cmidrule(lr){2-4} \cmidrule(lr){5-7} \cmidrule(lr){8-10}
\texttt{Vicuna} \texttt{Base} & 87.23 &  84.03 &      81.40 &     81.10 &      74.17 &   70.80 &     \textbf{78.80} &      \textbf{74.53} &   64.10 \\
\texttt{Vicuna} \texttt{+GC} & 85.73 &  83.93 &      81.43 &     83.17 &      76.20 &   73.17 &     73.57 &      69.27 &   65.73 \\
   \texttt{Vicuna} \texttt{+RE} & 85.30 &  84.07 &      81.90 &     79.83 &      \textbf{76.40} &   \textbf{74.67} &     76.00 &      71.43 &   65.83 \\
\texttt{Vicuna} \texttt{+RE+GC} & \textbf{88.27} &  \textbf{86.10} &      \textbf{82.37} &     \textbf{84.37} &      75.77 &   73.13 &     77.93 &      72.53 &   \textbf{65.80} \\
\cmidrule(lr){1-1} \cmidrule(lr){2-4} \cmidrule(lr){5-7} \cmidrule(lr){8-10}
   \texttt{Vicuna (Perc)} \texttt{Base} & \textbf{85.90} & \textbf{82.73} &      78.00 &     \textbf{79.37} &      69.40 &   65.10 &     \textbf{76.60} &      67.27 &   57.80 \\
   \texttt{Vicuna (Perc)} \texttt{+GC} & 83.93 &  82.23 &      78.33 &     76.37 &      69.77 &   64.97 &     73.20 &      66.47 &   59.20 \\
      \texttt{Vicuna (Perc)} \texttt{+RE} & 83.63 &  82.60 &      \textbf{78.37} &     76.40 &      \textbf{73.13} &   \textbf{70.03} &     69.13 &      \textbf{68.03} &   \textbf{62.33} \\
      \texttt{Vicuna (Perc)} \texttt{+RE+GC} & 84.97 &  80.27 &      76.03 &     78.20 &      71.30 &   67.90 &     71.87 &      65.90 &   60.27 \\
\bottomrule
\end{tabular}
}

    \caption{POPE results (accuracy) for MSCOCO, O365/COCO (using the 80 MSCOCO object classes), and O365/non-COCO (remaining 285 classes) for random, popular, and adversarial example sets.}
    \label{table:experiments:pope}
    \vspace{-1em}
\end{table*}

\begin{table}[]
    \centering
     \def\arraystretch{0.8}
     \resizebox{0.99\linewidth}{!}{
\begin{tabular}{lrrr}
\toprule
 \bf Model &     \bf R+ &  \bf  Rg &  \bf  R \\
\cmidrule(lr){1-1} \cmidrule(lr){2-4}
\texttt{Llama-3} \texttt{+RE} & 60.02 &     53.69 &    65.41 \\
\texttt{Llama-3} \texttt{+RE+GC} & 64.62 &     60.51 &    71.50 \\
\cmidrule(lr){1-1} \cmidrule(lr){2-4}
  \texttt{Phi-3} \texttt{+RE} & 63.33 &     61.06 &    67.09 \\
  \texttt{Phi-3} \texttt{+RE+GC} & 68.23 &     65.50 &    73.33 \\
\cmidrule(lr){1-1} \cmidrule(lr){2-4}
   \texttt{Vicuna} \texttt{+RE} & 58.03 &     58.78 &    61.89 \\
   \texttt{Vicuna} \texttt{+RE+GC} & 68.25 &     65.30 &    73.66 \\
\cmidrule(lr){1-1} \cmidrule(lr){2-4}
      \texttt{Vicuna (Perc)} \texttt{+RE} & 23.00 &     22.21 &    30.60 \\
      \texttt{Vicuna (Perc)} \texttt{+RE+GC} & 35.68 &     34.32 &    42.20 \\
\bottomrule
\end{tabular}
}
    \caption{Precision@50 for expression grounding (provide the bounding box for a region) for the test split of RefCOCO (R), RefCOCO+ (R+), and RefCOCOg (Rg). }
    \label{tab:ref_exp}
    \vspace{-1.3em}
\end{table}

\begin{table*}[t]
    \centering
    \footnotesize
     \def\arraystretch{0.8}
     \resizebox{0.99\linewidth}{!}{
\begin{tabular}{ll rrrrrrrr}
\toprule
    &     \bf  Model & \bf  CIDEr$\uparrow$ & \bf  CLIPS.$\uparrow$  & \bf  \#Words &  \bf CHAIR$_i\downarrow$  &  \bf Coverage$\uparrow$  &  \bf Objects &  \bf FaithScore$\uparrow$  &  \bf Facts \\
\cmidrule(lr){2-2} \cmidrule(lr){3-4} \cmidrule(lr){5-5} \cmidrule(lr){6-8} \cmidrule(lr){9-10}
\multirow{16}{*}{\bf MSCOCO} & \texttt{Llama-3} \texttt{Base} & 112.31 &11.71 &     10.22 &    3.84 &   56.43 &   1.61 &    91.25 &4.49 \\
 &     \texttt{Llama-3} \texttt{+GC} & 110.40 &11.33 &     10.68 &    3.61 &   54.34 &   1.56 &    90.74 &4.50 \\
 & \texttt{Llama-3} \texttt{+RE} & 109.01 &11.36 &     10.52 &    3.78 &   55.74 &   1.60 &    90.86 &4.64 \\
 & \texttt{Llama-3} \texttt{+RE+GC} & 107.95 &11.72 &     10.66 &    3.63 &   55.46 &   1.61 &    90.64 &4.69 \\
  \cmidrule(lr){2-2} \cmidrule(lr){3-4} \cmidrule(lr){5-5} \cmidrule(lr){6-8} \cmidrule(lr){9-10}
 &       \texttt{Phi-3} \texttt{Base} & 112.54 &11.97 &     11.41 &    3.28 &   57.54 &   1.68 &    90.98 &4.88 \\
 &       \texttt{Phi-3} \texttt{+GC} & 114.78 &12.15 &     11.06 &    3.83 &   56.55 &   1.66 &    90.90 &4.79 \\
 &   \texttt{Phi-3} \texttt{+RE} & 113.22 &12.07 &     11.14 &    3.43 &   57.18 &   1.68 &    91.06 &4.87 \\
 &   \texttt{Phi-3} \texttt{+RE+GC} & 113.68 &11.90 &     11.06 &    3.68 &   56.21 &   1.64 &    91.28 &4.66 \\
  \cmidrule(lr){2-2} \cmidrule(lr){3-4} \cmidrule(lr){5-5} \cmidrule(lr){6-8} \cmidrule(lr){9-10}
 & \texttt{Vicuna} \texttt{Base} & 115.57 &11.93 &     10.31 &    3.68 &   54.14 &   1.56 &    91.95 &4.61 \\
 & \texttt{Vicuna} \texttt{+GC} & 117.35 &11.80 &      9.82 &    3.08 &   53.98 &   1.50 &    92.05 &4.37 \\
 &    \texttt{Vicuna} \texttt{+RE} & 112.06 &11.76 &      9.92 &    3.41 &   54.21 &   1.55 &    92.19 &4.53 \\
 &    \texttt{Vicuna} \texttt{+RE+GC} & 113.30 &11.77 &      9.79 &    3.64 &   52.69 &   1.50 &    91.98 &4.27 \\
  \cmidrule(lr){2-2} \cmidrule(lr){3-4} \cmidrule(lr){5-5} \cmidrule(lr){6-8} \cmidrule(lr){9-10}
 &    \texttt{Vicuna (Perc)} \texttt{Base} & 107.74 &11.27 &     10.05 &    4.73 &   53.71 &   1.55 &    90.56 &4.46 \\
 &    \texttt{Vicuna (Perc)} \texttt{+GC} & 110.61 &11.50 &      9.86 &    4.16 &   54.11 &   1.53 &    90.53 &4.35 \\
 &       \texttt{Vicuna (Perc)} \texttt{+RE} & 107.38 &11.31 &      9.96 &    4.54 &   54.21 &   1.57 &    90.66 &4.51 \\
 &       \texttt{Vicuna (Perc)} \texttt{+RE+GC} & 109.64 &11.25 &     10.11 &    5.15 &   54.20 &   1.57 &    90.39 &4.56 \\
  \midrule
\multirow{16}{*}{\bf Objects365} &     \texttt{Llama-3} \texttt{Base} & --- &10.99 &     10.15 &   14.51 &   27.67 &   1.94 &    88.68 &4.56 \\
 &     \texttt{Llama-3} \texttt{+GC} & --- &10.84 &     10.72 &   13.33 &   26.72 &   1.84 &    88.88 &4.52 \\
 & \texttt{Llama-3} \texttt{+RE} & --- &10.67 &     10.50 &   12.74 &   26.73 &   1.86 &    88.57 &4.66 \\
 & \texttt{Llama-3} \texttt{+RE+GC} & --- &10.98 &     10.74 &   12.48 &   28.16 &   1.96 &    87.97 &4.86 \\
  \cmidrule(lr){2-2} \cmidrule(lr){3-4} \cmidrule(lr){5-5} \cmidrule(lr){6-8} \cmidrule(lr){9-10}
 &       \texttt{Phi-3} \texttt{Base} & --- &11.27 &     11.36 &   12.99 &   29.23 &   2.03 &    88.33 &4.77 \\
 &       \texttt{Phi-3} \texttt{+GC} & --- &11.60 &     11.08 &   13.17 &   28.73 &   1.96 &    88.90 &4.70 \\
 &   \texttt{Phi-3} \texttt{+RE} & --- &11.41 &     11.22 &   13.30 &   28.20 &   1.97 &    89.06 &4.88 \\
 &   \texttt{Phi-3} \texttt{+RE+GC} & --- &11.31 &     11.18 &   12.27 &   28.78 &   1.97 &    88.93 &4.64 \\
  \cmidrule(lr){2-2} \cmidrule(lr){3-4} \cmidrule(lr){5-5} \cmidrule(lr){6-8} \cmidrule(lr){9-10}
 & \texttt{Vicuna} \texttt{Base} & --- &11.06 &     10.28 &   12.44 &   27.38 &   1.88 &    88.81 &4.55 \\
 & \texttt{Vicuna} \texttt{+GC} & --- &11.12 &      9.78 &   12.62 &   26.23 &   1.76 &    89.82 &4.24 \\
 &    \texttt{Vicuna} \texttt{+RE} & --- &10.93 &     10.17 &   12.85 &   26.96 &   1.84 &    89.33 &4.58 \\
 &    \texttt{Vicuna} \texttt{+RE+GC} & --- &11.07 &      9.83 &   12.60 &   26.25 &   1.79 &    90.20 &4.24 \\
  \cmidrule(lr){2-2} \cmidrule(lr){3-4} \cmidrule(lr){5-5} \cmidrule(lr){6-8} \cmidrule(lr){9-10}
 &    \texttt{Vicuna (Perc.)} \texttt{Base} & --- &10.14 &     10.12 &   15.82 &   25.82 &   1.87 &    86.18 &4.36 \\
 &    \texttt{Vicuna (Perc)} \texttt{+GC} & --- &10.52 &      9.81 &   14.42 &   25.50 &   1.74 &    87.65 &4.19 \\
 &       \texttt{Vicuna (Perc)} \texttt{+RE} & --- &10.24 &     10.26 &   15.81 &   25.98 &   1.88 &    86.07 &4.55 \\
 &       \texttt{Vicuna (Perc)} \texttt{+RE+GC} & --- &10.30 &     10.23 &   16.68 &   25.92 &   1.84 &    86.50 &4.48 \\
\bottomrule
\end{tabular}
}
    \caption{Results on standard image captioning. CIDEr and CLIPScore indicate general caption quality; \texttt{CHAIR}$_i$ and \texttt{FaithScore} reflect hallucination, whereas (average number of) \#Words, CHAIR Coverage and Objects, and (number of FaithScore) Facts aim to quantify informativeness. 
    }
    \label{table:experiments:standard}
    \vspace{-1em}
\end{table*}

\begin{table*}[t]
    \centering
    \footnotesize
     \def\arraystretch{0.9}
     \resizebox{0.99\linewidth}{!}{
\begin{tabular}{ll rrrrrrrr}
\toprule
    &     \bf  Model & \bf  CIDEr$\uparrow$ & \bf  CLIPS.$\uparrow$  & \bf  \#Words &  \bf CHAIR$_i\downarrow$  &  \bf Coverage$\uparrow$  &  \bf Objects &  \bf FaithScore$\uparrow$  &  \bf Facts \\
\cmidrule(lr){2-2} \cmidrule(lr){3-4} \cmidrule(lr){5-5} \cmidrule(lr){6-8} \cmidrule(lr){9-10}
\multirow{8}{*}{\bf MSCOCO} &     \texttt{Llama-3} \texttt{+GC} &  -8.52 & 0.28 &     -0.48 &    0.17 &   -5.63 &  -0.21 &     1.12 &      -0.18 \\
& \texttt{Llama-3} \texttt{+RE+GC} &  -7.92 &-0.20 &     -0.44 &   -0.39 &   -5.44 &  -0.25 &     0.88 &      -0.28 \\
 \cmidrule(lr){2-2} \cmidrule(lr){3-4} \cmidrule(lr){5-5} \cmidrule(lr){6-8} \cmidrule(lr){9-10}
&       \texttt{Phi-3} \texttt{+GC} &  -6.23 &-0.25 &     -0.34 &   -0.14 &   -6.33 &  -0.28 &     0.63 &      -0.41 \\
&   \texttt{Phi-3} \texttt{+RE+GC} &  -8.12 &-0.17 &     -0.24 &    0.44 &   -7.36 &  -0.28 &     1.08 &      -0.29 \\
 \cmidrule(lr){2-2} \cmidrule(lr){3-4} \cmidrule(lr){5-5} \cmidrule(lr){6-8} \cmidrule(lr){9-10}
& \texttt{Vicuna} \texttt{+GC} &  -9.32 &-0.03 &      0.46 &    0.51 &   -6.64 &  -0.19 &     0.72 &      -0.09 \\
&    \texttt{Vicuna} \texttt{+RE+GC} &  -8.22 & 0.09 &      0.91 &    0.03 &   -4.80 &  -0.19 &     0.48 &0.11 \\
 \cmidrule(lr){2-2} \cmidrule(lr){3-4} \cmidrule(lr){5-5} \cmidrule(lr){6-8} \cmidrule(lr){9-10}
&    \texttt{Vicuna (Perc.)} \texttt{+GC} &  -7.78 &-0.22 &      0.12 &    0.06 &   -6.87 &  -0.22 &     0.61 &      -0.18 \\
&       \texttt{Vicuna (Perc.)} \texttt{+RE+GC} & -13.69 &-0.16 &      0.23 &   -1.08 &   -8.13 &  -0.32 &     0.87 &      -0.19 \\
 \midrule
\multirow{8}{*}{\bf Objects365} & \texttt{Llama-3} \texttt{+GC} &---&-0.02 &     -0.50 &   -1.07 &   -3.06 &  -0.25 &     0.46 &      -0.18 \\
& \texttt{Llama-3} \texttt{+RE+GC} &---&-0.34 &     -0.31 &   -0.01 &   -3.67 &  -0.32 &     1.09 &      -0.30 \\
 \cmidrule(lr){2-2} \cmidrule(lr){3-4} \cmidrule(lr){5-5} \cmidrule(lr){6-8} \cmidrule(lr){9-10}
&       \texttt{Phi-3} \texttt{+GC} &---&-0.39 &     -0.03 &   -1.91 &   -2.89 &  -0.26 &     0.87 &      -0.22 \\
&   \texttt{Phi-3} \texttt{+RE+GC} &---&-0.28 &     -0.05 &   -0.48 &   -3.12 &  -0.28 &     0.74 &      -0.09 \\
 \cmidrule(lr){2-2} \cmidrule(lr){3-4} \cmidrule(lr){5-5} \cmidrule(lr){6-8} \cmidrule(lr){9-10}
& \texttt{Vicuna} \texttt{+GC} &---& 0.04 &      0.44 &   -1.38 &   -2.03 &  -0.17 &     0.21 &0.09 \\
&    \texttt{Vicuna} \texttt{+RE+GC} &---&-0.06 &      0.86 &   -1.06 &   -3.35 &  -0.27 &    -0.25 &0.26 \\
 \cmidrule(lr){2-2} \cmidrule(lr){3-4} \cmidrule(lr){5-5} \cmidrule(lr){6-8} \cmidrule(lr){9-10}
&    \texttt{Vicuna (Perc.)} \texttt{+GC} &---&-0.00 &      0.25 &   -0.77 &   -2.61 &  -0.21 &    -0.14 &0.03 \\
&       \texttt{Vicuna (Perc.)} \texttt{+RE+GC} &---&-0.12 &      0.30 &   -2.37 &   -3.40 &  -0.37 &     1.59 &      -0.06 \\
\bottomrule
\end{tabular}
}
    \caption{Absolute performance difference of  grounded image captioning w.r.t. standard captioning (Table~\ref{table:experiments:standard}). 
    }
    \label{table:experiments:grounded}
\end{table*}

\section{Results}
\label{ssec:results}

We now report the observed hallucination effects under both protocols: in free-form captioning and in QA-based hallucination evaluation (as indicated by the POPE metric/protocol).
The reported CHAIR results correspond to our CHAIR-MEN variant; we report the results obtained with the vanilla CHAIR based on string matching in Appendix \ref{sec:appendix:chair}.
We did not separately optimize hyperparameters for each LLM and will thus refrain from their mutual performance comparison; instead, for each of the three LLMs, we analyze how inclusion of grounding objectives affects their hallucination.   

\rparagraph{Referring Expressions}
\label{sec:appendix:ref_ex_results}
Before we test the effects of grounding on free-form and QA-based hallucination, we first analyze if the two grounding objectives are mutually compatible. Concretely, we test how the models trained with grounding objectives (\texttt{+RE}, and \texttt{+RE+GC}) perform on one of the grounding tasks itself. In other words, we test if and how well models explicitly trained with grounding objectives learn to ground expressions and whether the two grounding objectives are mutually beneficial. 
The results for expression grounding (one of the two \texttt{RE} tasks: given the description, provide the bounding box) are shown in Table~\ref{tab:ref_exp}. The metric is precision@50, that is, the proportion of examples where the intersection between the predicted and gold bounding box contains at least 50\% of their union.
The results indicate that adding grounded captioning (\texttt{+GC}) consistently and substantially improves the performance for all three LLMs: this strongly suggests that the two grounding objectives are mutually compatible. Vicuna-based model with the perceiver-resampler (\texttt{Perc}) aligner considerably underperforms the (default) MLP aligner; we suspect that this is because the (pre-)training data was insufficient for it to learn to properly encode positional information.

\rparagraph{QA Hallucinations with POPE}
Table~\ref{table:experiments:pope} summarizes the hallucination results according to the QA-based evaluation protocol with POPE. 
Overall, both grounding objectives, referring expressions (\texttt{+RE}) and grounding captions (\texttt{+GC}) fail to consistently and non-negligibly improve performance, i.e., reduce hallucination. While their combination \texttt{+RE+GC} greatly improves grounding capabilities over \texttt{+RE} alone for all LLMs (Table \ref{tab:ref_exp}), the same is not true for QA-based hallucination reduction (i.e., POPE), pointing to the lack of causal link between object grounding and hallucination reduction.


\rparagraph{Standard Captions}
Table~\ref{table:experiments:standard} displays the performance of our LVLM variants on standard image captioning.
We observe consistently, for all tested models on both evaluation datasets, that grounding objectives (i.e., their inclusion or exclusion) have little to no effect on performance: all models learn to generate proper captions in the MSCOCO style, with 10 words on average and of similar general quality, as captured by the caption quality metrics (CIDEr, CLIPScore). The metrics that capture caption detailness (coverage, number of objects \& atomic facts) also show little difference between the models. Most importantly, the same is true for hallucination metrics CHAIR$_i$ and FaithScore, confirming that there is \textbf{no} positive transfer from grounding to hallucination reduction.

\rparagraph{Grounded Captions}
Previous results establish that \textit{training} on grounding objectives does not reduce hallucination in open caption generation. We next test whether forcing the model to generate grounded captions at \textit{inference} can reduce hallucination. Intuitively, prompting the model to produce grounded captions should encourage it to generate only objects contained in the image.
The results in Table~\ref{table:experiments:grounded} show that generating grounded captions indeed results in some hallucination reduction, but the effect is rather small. Reduction is more prominent on Objects365 where the baseline hallucination rate is higher than on MSCOCO.  
On the flip side, generating grounded captions at inference slightly reduces their informativeness too (i.e., we observe fewer objects and atomic facts in the generated captions).
A closer qualitative inspection (see \S\ref{sec:qualitative}) reveals that LVLMs trained with grounding objectives still incorrectly describe objects or fabricate them entirely.



\section{Qualitative Grounded Caption Analysis}
\label{sec:qualitative}

\begin{figure}[t]
\setulcolor{red} 
\footnotesize
 \begin{minipage}{0.2\textwidth}
 \includegraphics[width=\textwidth]{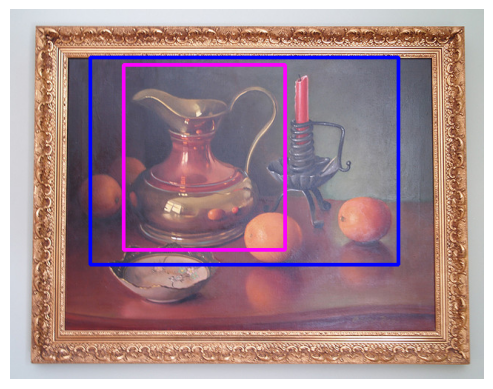} 
    \end{minipage}
    \hspace{0.05cm} 
    \begin{minipage}{0.26\textwidth}
\textit{Standard}: A painting of \ul{a woman} with a vase and oranges.  \\
\textit{Grounded}:
An artistic painting of \textcolor{blue}{\ul{a woman}}
with \textcolor{magenta}{a vase}
.  \\
    \end{minipage}
    
\begin{minipage}{0.2\textwidth}
    \includegraphics[width=\textwidth]{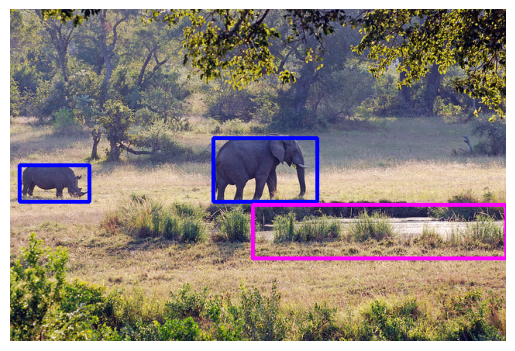} 
\end{minipage}
\hspace{0.05cm} 
\begin{minipage}{0.26\textwidth}
\textit{Standard}: \ul{Two elephants} are in a field near water.  \\
\textit{Grounded}: \textcolor{blue}{\ul{Two elephants}} are in a field with \textcolor{magenta}{water}. 
  \\
    \end{minipage}

 \begin{minipage}{0.2\textwidth}
 \includegraphics[width=\textwidth]{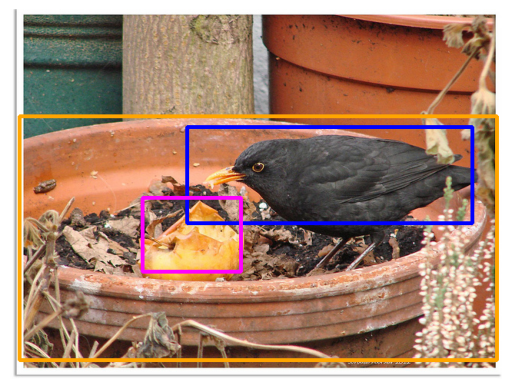} 
    \end{minipage}
    \hspace{0.05cm} 
    \begin{minipage}{0.26\textwidth}
\textit{Standard}:  A small bird is standing in a \ul{pot of food}.  \\
\textit{Grounded}:
\textcolor{blue}{A black bird} is eating \textcolor{magenta}{a \ul{peeled} apple} out of \textcolor{orange}{a pot}
.  \\
    \end{minipage}
    


    \caption{Qualitative examples of \texttt{Vicuna +RE+GC} for standard and grounded captioning. Hallucinations are underlined \ul{in red}.
    Predicted bounding boxes are visualized in the image and marked in the caption.
    }
    \label{figure:example_grounded}
    \vspace{-1em}
\end{figure}

We show examples for grounded captioning in Figure~\ref{figure:example_grounded}. 
The grounding itself does not necessarily prevent the model from hallucinating: 
in the first example, the model fully hallucinates a woman along with a bounding box for her.
In the second example, the second `elephant' bounding box is positionally correct in that it points to an animal, but that animal is a rhino.   
In the third example, similarly, the bounding box correctly contains an apple but the attribute `peeled' is hallucinated. These examples point to causes of hallucination that go beyond insufficient or incorrect grounding and help explain why grounding objectives do not really reduce the LVLM hallucination in open captioning.  





\section{Related Work}

\noindent\textbf{Large Vision-Language Models.} 
LVLMs are essentially Large Language Models (LLMs) \cite{brown_language_2020,touvron_llama_2023,openai_gpt-4_2023,jiang_mistral_2023} extended to ``understand'' visual input.
Recent models have shown an impressive understanding of images \cite{openai_gpt-4_2023,anil_gemini_2023,li_blip-2_2023,dai_instructblip_2023,liu_visual_2023,bai_qwen-vl_2023,fini_improved_2023,zhu_minigpt-4_2023,laurencon_obelisc_2023,geigle_mblip_2023,wang_cogvlm_2023} and a range of models have been proposed specifically for grounding and referring \cite{chen_shikra_2023,you_ferret_2023,pramanick_jack_2023,zhang_gpt4roi_2023,peng_kosmos-2_2023,chen_position-enhanced_2023,zhao_bubogpt_2023}.

\rparagraph{Measuring Object Hallucinations}
A range of hallucination metrics have been proposed: CHAIR \cite{rohrbach_object_2018} identifies hallucinated objects by checking captions (via string matching) against a set of annotated objects (i.e., MSCOCO).
\citet{wang_evaluation_2023} fine-tune an LLM to identify hallucinatory captions through comparison with reference captions; FaithScore \cite{jing_faithscore_2023}, a reference-free approach, uses an LLM to extract verifiable facts and then tests these facts with a VQA model. 
POPE \cite{li_evaluating_2023} indirectly measures hallucination with questions about object existence: while a good test of image understanding
, which may indicate the extent of models' tendency to hallucinate, it is not a direct measure of hallucination in open-ended captioning.

\paragraph{Hallucination Mitigation.} A range of approaches have been proposed to mitigate hallucination:
\citet{biten_let_2022,dai_plausible_2023,zhai_halle-switch_2023} propose adaptions to the training data and objectives. 
\citet{liu_aligning_2023,gunjal_detecting_2023,zhao_beyond_2023,yu_rlhf-v_2023} use reinforcement-learning methods to reduce hallucinations in model output.
\citet{leng_mitigating_2023,huang_opera_2023} propose (training-free)  decoding methods that mitigate hallucinations. 
\citet{zhou_analyzing_2023,yin_woodpecker_2023} create pipeline approaches that post-hoc clean the generated text from hallucinated content.
Finally, for QA hallucinations, researchers have created robust instruction data \cite{liu_aligning_2023}, VQA examples \cite{hu_ciem_2023}, and additional benchmarks \cite{lu_evaluation_2023}.

\section{Conclusion}
Object hallucination remains one of the main obstacles to wide-range adoption of LVLMs. Prior work suggested that grounding objectives like referring expressions reduce hallucination but the empirical support for this claim is confined to QA-based evaluation.
In this work, we carried out an in-depth analysis of the effects that grounding objectives in LVLM training have on their hallucination in open image captioning.
Our extensive experiments with three backbone LLMs show that there is \textit{no} causal link between improved object grounding (via objectives like referring expressions) and hallucination reduction: this observation is true both under QA-based and open captioning hallucination evaluation protocols. 
Finally, we observe that explicitly prompting LVLMs to generate grounded captions at inference can slightly reduce hallucination but at the expense of reduced caption informativeness.



\FloatBarrier

\newpage
\section{Limitations}
There are two main limitations to our analysis. First, while we aim for a comprehensive analysis of the effects of different training objectives and task mixes on downstream hallucination, there are a number of modeling decisions that we had to fix (i.e., we could not explore other variants)---primarily w.r.t. to the architecture of the LVLM--- due to a limited computational budget. One could, inter alia, consider a different image encoder, additional or larger LLMs, and/or alignment modules other than the MLP or perceiver-resampler.  
Additionally, due to our limited computational budget, we train our models on less data and for fewer steps than a lot of other work that trains LVLMs (e.g. \citet{chen_shikra_2023,liu_improved_2023,bai_qwen-vl_2023}); we thus cannot rule out that a reduction in hallucination due to grounding objectives might \textit{emerge} at some larger scale of grounding training.

Second, our findings are (modulo anecdotal evidence from manual qualitative analysis of a limited number of examples) based on reliance on imperfect automatic metrics. While this is a common practice in related work as well, we increase the likelihood of the robustness of our findings and conclusions by employing two mutually complementing hallucination quantification metrics, CHAIR and FaithScore (see \S\ref{sec:metrics}), as well as additionally proposing a semantic extension to CHAIR (CHAIR-MEN, see \S\ref{sec:metrics}). 


\section*{Acknowledgements}

This work was in part supported by the Alexander von Humboldt Foundation.

\bibliography{anthology,custom}

\begin{thebibliography}{69}
\expandafter\ifx\csname natexlab\endcsname\relax\def\natexlab#1{#1}\fi

\bibitem[{Abdin et~al.(2024)Abdin, Jacobs, Awan, Aneja, Awadallah, Awadalla, Bach, Bahree, Bakhtiari, Behl, Benhaim, Bilenko, Bjorck, Bubeck, Cai, Mendes, Chen, Chaudhary, Chopra, Giorno, Rosa, Dixon, Eldan, Iter, Garg, Goswami, Gunasekar, Haider, Hao, Hewett, Huynh, Javaheripi, Jin, Kauffmann, Karampatziakis, Kim, Khademi, Kurilenko, Lee, Lee, Li, Liang, Liu, Lin, Lin, Madan, Mitra, Modi, Nguyen, Norick, Patra, Perez-Becker, Portet, Pryzant, Qin, Radmilac, Rosset, Roy, Ruwase, Saarikivi, Saied, Salim, Santacroce, Shah, Shang, Sharma, Song, Tanaka, Wang, Ward, Wang, Witte, Wyatt, Xu, Xu, Yadav, Yang, Yang, Yu, Zhang, Zhang, Zhang, Zhang, Zhang, Zhang, Zhang, and Zhou}]{abdin_phi-3_2024}
Marah Abdin, Sam~Ade Jacobs, Ammar~Ahmad Awan, Jyoti Aneja, Ahmed Awadallah, Hany Awadalla, Nguyen Bach, Amit Bahree, Arash Bakhtiari, Harkirat Behl, Alon Benhaim, Misha Bilenko, Johan Bjorck, Sébastien Bubeck, Martin Cai, Caio César~Teodoro Mendes, Weizhu Chen, Vishrav Chaudhary, Parul Chopra, Allie~Del Giorno, Gustavo~de Rosa, Matthew Dixon, Ronen Eldan, Dan Iter, Amit Garg, Abhishek Goswami, Suriya Gunasekar, Emman Haider, Junheng Hao, Russell~J. Hewett, Jamie Huynh, Mojan Javaheripi, Xin Jin, Piero Kauffmann, Nikos Karampatziakis, Dongwoo Kim, Mahoud Khademi, Lev Kurilenko, James~R. Lee, Yin~Tat Lee, Yuanzhi Li, Chen Liang, Weishung Liu, Eric Lin, Zeqi Lin, Piyush Madan, Arindam Mitra, Hardik Modi, Anh Nguyen, Brandon Norick, Barun Patra, Daniel Perez-Becker, Thomas Portet, Reid Pryzant, Heyang Qin, Marko Radmilac, Corby Rosset, Sambudha Roy, Olatunji Ruwase, Olli Saarikivi, Amin Saied, Adil Salim, Michael Santacroce, Shital Shah, Ning Shang, Hiteshi Sharma, Xia Song, Masahiro Tanaka, Xin Wang, Rachel
  Ward, Guanhua Wang, Philipp Witte, Michael Wyatt, Can Xu, Jiahang Xu, Sonali Yadav, Fan Yang, Ziyi Yang, Donghan Yu, Chengruidong Zhang, Cyril Zhang, Jianwen Zhang, Li~Lyna Zhang, Yi~Zhang, Yue Zhang, Yunan Zhang, and Xiren Zhou. 2024.
\newblock Phi-3 {Technical} {Report}: {A} {Highly} {Capable} {Language} {Model} {Locally} on {Your} {Phone}.
\newblock \_eprint: 2404.14219.

\bibitem[{{AI@Meta}(2024)}]{aimeta_llama_2024}
{AI@Meta}. 2024.
\newblock \href {https://github.com/meta-llama/llama3/blob/main/MODEL_CARD.md} {Llama 3 {Model} {Card}}.

\bibitem[{Alayrac et~al.(2022)Alayrac, Donahue, Luc, Miech, Barr, Hasson, Lenc, Mensch, Millican, Reynolds, Ring, Rutherford, Cabi, Han, Gong, Samangooei, Monteiro, Menick, Borgeaud, Brock, Nematzadeh, Sharifzadeh, Binkowski, Barreira, Vinyals, Zisserman, and Simonyan}]{alayrac_flamingo_2022}
Jean-Baptiste Alayrac, Jeff Donahue, Pauline Luc, Antoine Miech, Iain Barr, Yana Hasson, Karel Lenc, Arthur Mensch, Katie Millican, Malcolm Reynolds, Roman Ring, Eliza Rutherford, Serkan Cabi, Tengda Han, Zhitao Gong, Sina Samangooei, Marianne Monteiro, Jacob Menick, Sebastian Borgeaud, Andrew Brock, Aida Nematzadeh, Sahand Sharifzadeh, Mikolaj Binkowski, Ricardo Barreira, Oriol Vinyals, Andrew Zisserman, and Karen Simonyan. 2022.
\newblock \href {https://doi.org/10.48550/arXiv.2204.14198} {Flamingo: a {Visual} {Language} {Model} for {Few}-{Shot} {Learning}}.
\newblock \emph{CoRR}, abs/2204.14198.
\newblock ArXiv: 2204.14198.

\bibitem[{Anil et~al.(2023)Anil, Borgeaud, Wu, Alayrac, Yu, Soricut, Schalkwyk, Dai, Hauth, Millican, Silver, Petrov, Johnson, Antonoglou, Schrittwieser, Glaese, Chen, Pitler, Lillicrap, Lazaridou, Firat, Molloy, Isard, Barham, Hennigan, Lee, Viola, Reynolds, Xu, Doherty, Collins, Meyer, Rutherford, Moreira, Ayoub, Goel, Tucker, Piqueras, Krikun, Barr, Savinov, Danihelka, Roelofs, White, Andreassen, Glehn, Yagati, Kazemi, Gonzalez, Khalman, Sygnowski, and al}]{anil_gemini_2023}
Rohan Anil, Sebastian Borgeaud, Yonghui Wu, Jean-Baptiste Alayrac, Jiahui Yu, Radu Soricut, Johan Schalkwyk, Andrew~M. Dai, Anja Hauth, Katie Millican, David Silver, Slav Petrov, Melvin Johnson, Ioannis Antonoglou, Julian Schrittwieser, Amelia Glaese, Jilin Chen, Emily Pitler, Timothy~P. Lillicrap, Angeliki Lazaridou, Orhan Firat, James Molloy, Michael Isard, Paul~Ronald Barham, Tom Hennigan, Benjamin Lee, Fabio Viola, Malcolm Reynolds, Yuanzhong Xu, Ryan Doherty, Eli Collins, Clemens Meyer, Eliza Rutherford, Erica Moreira, Kareem Ayoub, Megha Goel, George Tucker, Enrique Piqueras, Maxim Krikun, Iain Barr, Nikolay Savinov, Ivo Danihelka, Becca Roelofs, Anaïs White, Anders Andreassen, Tamara~von Glehn, Lakshman Yagati, Mehran Kazemi, Lucas Gonzalez, Misha Khalman, Jakub Sygnowski, and et~al. 2023.
\newblock \href {https://doi.org/10.48550/ARXIV.2312.11805} {Gemini: {A} {Family} of {Highly} {Capable} {Multimodal} {Models}}.
\newblock \emph{CoRR}, abs/2312.11805.
\newblock ArXiv: 2312.11805.

\bibitem[{Bai et~al.(2023)Bai, Bai, Yang, Wang, Tan, Wang, Lin, Zhou, and Zhou}]{bai_qwen-vl_2023}
Jinze Bai, Shuai Bai, Shusheng Yang, Shijie Wang, Sinan Tan, Peng Wang, Junyang Lin, Chang Zhou, and Jingren Zhou. 2023.
\newblock \href {https://doi.org/10.48550/ARXIV.2308.12966} {Qwen-{VL}: {A} {Frontier} {Large} {Vision}-{Language} {Model} with {Versatile} {Abilities}}.
\newblock \emph{CoRR}, abs/2308.12966.
\newblock ArXiv: 2308.12966.

\bibitem[{Biten et~al.(2022)Biten, Gómez, and Karatzas}]{biten_let_2022}
Ali~Furkan Biten, Lluís Gómez, and Dimosthenis Karatzas. 2022.
\newblock \href {https://doi.org/10.1109/WACV51458.2022.00253} {Let there be a clock on the beach: {Reducing} {Object} {Hallucination} in {Image} {Captioning}}.
\newblock In \emph{{IEEE}/{CVF} {Winter} {Conference} on {Applications} of {Computer} {Vision}, {WACV} 2022, {Waikoloa}, {HI}, {USA}, {January} 3-8, 2022}, pages 2473--2482. IEEE.

\bibitem[{Brown et~al.(2020)Brown, Mann, Ryder, Subbiah, Kaplan, Dhariwal, Neelakantan, Shyam, Sastry, Askell, Agarwal, Herbert-Voss, Krueger, Henighan, Child, Ramesh, Ziegler, Wu, Winter, Hesse, Chen, Sigler, Litwin, Gray, Chess, Clark, Berner, McCandlish, Radford, Sutskever, and Amodei}]{brown_language_2020}
Tom~B. Brown, Benjamin Mann, Nick Ryder, Melanie Subbiah, Jared Kaplan, Prafulla Dhariwal, Arvind Neelakantan, Pranav Shyam, Girish Sastry, Amanda Askell, Sandhini Agarwal, Ariel Herbert-Voss, Gretchen Krueger, Tom Henighan, Rewon Child, Aditya Ramesh, Daniel~M. Ziegler, Jeffrey Wu, Clemens Winter, Christopher Hesse, Mark Chen, Eric Sigler, Mateusz Litwin, Scott Gray, Benjamin Chess, Jack Clark, Christopher Berner, Sam McCandlish, Alec Radford, Ilya Sutskever, and Dario Amodei. 2020.
\newblock \href {http://arxiv.org/abs/2005.14165} {Language {Models} are {Few}-{Shot} {Learners}}.
\newblock \emph{arXiv:2005.14165 [cs]}.
\newblock ArXiv: 2005.14165.

\bibitem[{Bugliarello et~al.(2023)Bugliarello, Sartran, Agrawal, Hendricks, and Nematzadeh}]{bugliarello_measuring_2023-1}
Emanuele Bugliarello, Laurent Sartran, Aishwarya Agrawal, Lisa~Anne Hendricks, and Aida Nematzadeh. 2023.
\newblock \href {https://doi.org/10.18653/V1/2023.ACL-LONG.87} {Measuring {Progress} in {Fine}-grained {Vision}-and-{Language} {Understanding}}.
\newblock In \emph{Proceedings of the 61st {Annual} {Meeting} of the {Association} for {Computational} {Linguistics} ({Volume} 1: {Long} {Papers}), {ACL} 2023, {Toronto}, {Canada}, {July} 9-14, 2023}, pages 1559--1582. Association for Computational Linguistics.

\bibitem[{Chen et~al.(2023{\natexlab{a}})Chen, Qin, Luo, Mi, Li, Sun, and Liu}]{chen_position-enhanced_2023}
Chi Chen, Ruoyu Qin, Fuwen Luo, Xiaoyue Mi, Peng Li, Maosong Sun, and Yang Liu. 2023{\natexlab{a}}.
\newblock \href {https://doi.org/10.48550/ARXIV.2308.13437} {Position-{Enhanced} {Visual} {Instruction} {Tuning} for {Multimodal} {Large} {Language} {Models}}.
\newblock \emph{CoRR}, abs/2308.13437.
\newblock ArXiv: 2308.13437.

\bibitem[{Chen et~al.(2023{\natexlab{b}})Chen, Zhang, Zeng, Zhang, Zhu, and Zhao}]{chen_shikra_2023}
Keqin Chen, Zhao Zhang, Weili Zeng, Richong Zhang, Feng Zhu, and Rui Zhao. 2023{\natexlab{b}}.
\newblock \href {https://doi.org/10.48550/ARXIV.2306.15195} {Shikra: {Unleashing} {Multimodal} {LLM}'s {Referential} {Dialogue} {Magic}}.
\newblock \emph{CoRR}, abs/2306.15195.
\newblock ArXiv: 2306.15195.

\bibitem[{Chiang et~al.(2023)Chiang, Li, Lin, Sheng, Wu, Zhang, Zheng, Zhuang, Zhuang, Gonzalez, Stoica, and Xing}]{chiang_vicuna_2023}
Wei-Lin Chiang, Zhuohan Li, Zi~Lin, Ying Sheng, Zhanghao Wu, Hao Zhang, Lianmin Zheng, Siyuan Zhuang, Yonghao Zhuang, Joseph~E. Gonzalez, Ion Stoica, and Eric~P. Xing. 2023.
\newblock \href {https://lmsys.org/blog/2023-03-30-vicuna/} {Vicuna: {An} {Open}-{Source} {Chatbot} {Impressing} {GPT}-4 with 90\%* {ChatGPT} {Quality}}.

\bibitem[{Chu et~al.(2024)Chu, Qiao, Zhang, Xu, Wei, Yang, Sun, Hu, Lin, Zhang, and Shen}]{chu_mobilevlm_2024}
Xiangxiang Chu, Limeng Qiao, Xinyu Zhang, Shuang Xu, Fei Wei, Yang Yang, Xiaofei Sun, Yiming Hu, Xinyang Lin, Bo~Zhang, and Chunhua Shen. 2024.
\newblock \href {https://doi.org/10.48550/ARXIV.2402.03766} {{MobileVLM} {V2}: {Faster} and {Stronger} {Baseline} for {Vision} {Language} {Model}}.
\newblock \emph{CoRR}, abs/2402.03766.
\newblock ArXiv: 2402.03766.

\bibitem[{Dai et~al.(2023{\natexlab{a}})Dai, Li, Li, Tiong, Zhao, Wang, Li, Fung, and Hoi}]{dai_instructblip_2023}
Wenliang Dai, Junnan Li, Dongxu Li, Anthony Meng~Huat Tiong, Junqi Zhao, Weisheng Wang, Boyang Li, Pascale Fung, and Steven C.~H. Hoi. 2023{\natexlab{a}}.
\newblock \href {https://doi.org/10.48550/arXiv.2305.06500} {{InstructBLIP}: {Towards} {General}-purpose {Vision}-{Language} {Models} with {Instruction} {Tuning}}.
\newblock \emph{CoRR}, abs/2305.06500.
\newblock ArXiv: 2305.06500.

\bibitem[{Dai et~al.(2023{\natexlab{b}})Dai, Liu, Ji, Su, and Fung}]{dai_plausible_2023}
Wenliang Dai, Zihan Liu, Ziwei Ji, Dan Su, and Pascale Fung. 2023{\natexlab{b}}.
\newblock \href {https://doi.org/10.18653/V1/2023.EACL-MAIN.156} {Plausible {May} {Not} {Be} {Faithful}: {Probing} {Object} {Hallucination} in {Vision}-{Language} {Pre}-training}.
\newblock In \emph{Proceedings of the 17th {Conference} of the {European} {Chapter} of the {Association} for {Computational} {Linguistics}, {EACL} 2023, {Dubrovnik}, {Croatia}, {May} 2-6, 2023}, pages 2128--2140. Association for Computational Linguistics.

\bibitem[{Dettmers et~al.(2023)Dettmers, Pagnoni, Holtzman, and Zettlemoyer}]{dettmers_qlora_2023-1}
Tim Dettmers, Artidoro Pagnoni, Ari Holtzman, and Luke Zettlemoyer. 2023.
\newblock \href {https://doi.org/10.48550/arXiv.2305.14314} {{QLoRA}: {Efficient} {Finetuning} of {Quantized} {LLMs}}.
\newblock \emph{CoRR}, abs/2305.14314.

\bibitem[{Fini et~al.(2023)Fini, Astolfi, Romero-Soriano, Verbeek, and Drozdzal}]{fini_improved_2023}
Enrico Fini, Pietro Astolfi, Adriana Romero-Soriano, Jakob Verbeek, and Michal Drozdzal. 2023.
\newblock \href {https://doi.org/10.48550/ARXIV.2305.08675} {Improved baselines for vision-language pre-training}.
\newblock \emph{CoRR}, abs/2305.08675.
\newblock ArXiv: 2305.08675.

\bibitem[{Geigle et~al.(2023)Geigle, Jain, Timofte, and Glavas}]{geigle_mblip_2023}
Gregor Geigle, Abhay Jain, Radu Timofte, and Goran Glavas. 2023.
\newblock \href {https://doi.org/10.48550/ARXIV.2307.06930} {{mBLIP}: {Efficient} {Bootstrapping} of {Multilingual} {Vision}-{LLMs}}.
\newblock \emph{CoRR}, abs/2307.06930.
\newblock ArXiv: 2307.06930.

\bibitem[{Goyal et~al.(2017)Goyal, Khot, Summers-Stay, Batra, and Parikh}]{goyal_making_2017}
Y.~Goyal, T.~Khot, D.~Summers-Stay, D.~Batra, and D.~Parikh. 2017.
\newblock \href {https://doi.org/10.1109/CVPR.2017.670} {Making the {V} in {VQA} {Matter}: {Elevating} the {Role} of {Image} {Understanding} in {Visual} {Question} {Answering}}.
\newblock In \emph{2017 {IEEE} {Conference} on {Computer} {Vision} and {Pattern} {Recognition} ({CVPR})}, pages 6325--6334.

\bibitem[{Gunjal et~al.(2023)Gunjal, Yin, and Bas}]{gunjal_detecting_2023}
Anisha Gunjal, Jihan Yin, and Erhan Bas. 2023.
\newblock \href {https://doi.org/10.48550/ARXIV.2308.06394} {Detecting and {Preventing} {Hallucinations} in {Large} {Vision} {Language} {Models}}.
\newblock \emph{CoRR}, abs/2308.06394.
\newblock ArXiv: 2308.06394.

\bibitem[{Gupta et~al.(2019)Gupta, Dollár, and Girshick}]{gupta_lvis_2019}
Agrim Gupta, Piotr Dollár, and Ross~B. Girshick. 2019.
\newblock \href {https://doi.org/10.1109/CVPR.2019.00550} {{LVIS}: {A} {Dataset} for {Large} {Vocabulary} {Instance} {Segmentation}}.
\newblock In \emph{{IEEE} {Conference} on {Computer} {Vision} and {Pattern} {Recognition}, {CVPR} 2019, {Long} {Beach}, {CA}, {USA}, {June} 16-20, 2019}, pages 5356--5364. Computer Vision Foundation / IEEE.

\bibitem[{Hu et~al.(2022)Hu, Shen, Wallis, Allen-Zhu, Li, Wang, Wang, and Chen}]{hu_lora_2022-2}
Edward~J. Hu, Yelong Shen, Phillip Wallis, Zeyuan Allen-Zhu, Yuanzhi Li, Shean Wang, Lu~Wang, and Weizhu Chen. 2022.
\newblock \href {https://openreview.net/forum?id=nZeVKeeFYf9} {{LoRA}: {Low}-{Rank} {Adaptation} of {Large} {Language} {Models}}.
\newblock In \emph{The {Tenth} {International} {Conference} on {Learning} {Representations}, {ICLR} 2022, {Virtual} {Event}, {April} 25-29, 2022}. OpenReview.net.

\bibitem[{Hu et~al.(2023)Hu, Zhang, Zhao, and Sun}]{hu_ciem_2023}
Hongyu Hu, Jiyuan Zhang, Minyi Zhao, and Zhenbang Sun. 2023.
\newblock \href {https://doi.org/10.48550/ARXIV.2309.02301} {{CIEM}: {Contrastive} {Instruction} {Evaluation} {Method} for {Better} {Instruction} {Tuning}}.
\newblock \emph{CoRR}, abs/2309.02301.
\newblock ArXiv: 2309.02301.

\bibitem[{Huang et~al.(2023)Huang, Dong, Zhang, Wang, He, Wang, Lin, Zhang, and Yu}]{huang_opera_2023}
Qidong Huang, Xiaoyi Dong, Pan Zhang, Bin Wang, Conghui He, Jiaqi Wang, Dahua Lin, Weiming Zhang, and Nenghai Yu. 2023.
\newblock \href {https://doi.org/10.48550/ARXIV.2311.17911} {{OPERA}: {Alleviating} {Hallucination} in {Multi}-{Modal} {Large} {Language} {Models} via {Over}-{Trust} {Penalty} and {Retrospection}-{Allocation}}.
\newblock \emph{CoRR}, abs/2311.17911.
\newblock ArXiv: 2311.17911.

\bibitem[{Jiang et~al.(2023)Jiang, Sablayrolles, Mensch, Bamford, Chaplot, Casas, Bressand, Lengyel, Lample, Saulnier, Lavaud, Lachaux, Stock, Scao, Lavril, Wang, Lacroix, and Sayed}]{jiang_mistral_2023}
Albert~Q. Jiang, Alexandre Sablayrolles, Arthur Mensch, Chris Bamford, Devendra~Singh Chaplot, Diego de~Las Casas, Florian Bressand, Gianna Lengyel, Guillaume Lample, Lucile Saulnier, Lélio~Renard Lavaud, Marie-Anne Lachaux, Pierre Stock, Teven~Le Scao, Thibaut Lavril, Thomas Wang, Timothée Lacroix, and William~El Sayed. 2023.
\newblock \href {https://doi.org/10.48550/ARXIV.2310.06825} {Mistral {7B}}.
\newblock \emph{CoRR}, abs/2310.06825.
\newblock ArXiv: 2310.06825.

\bibitem[{Jing et~al.(2023)Jing, Li, Chen, Jia, and Du}]{jing_faithscore_2023}
Liqiang Jing, Ruosen Li, Yunmo Chen, Mengzhao Jia, and Xinya Du. 2023.
\newblock \href {https://doi.org/10.48550/ARXIV.2311.01477} {{FAITHSCORE}: {Evaluating} {Hallucinations} in {Large} {Vision}-{Language} {Models}}.
\newblock \emph{CoRR}, abs/2311.01477.
\newblock ArXiv: 2311.01477.

\bibitem[{Kazemzadeh et~al.(2014)Kazemzadeh, Ordonez, Matten, and Berg}]{kazemzadeh_referitgame_2014}
Sahar Kazemzadeh, Vicente Ordonez, Mark Matten, and Tamara~L. Berg. 2014.
\newblock \href {https://doi.org/10.3115/V1/D14-1086} {{ReferItGame}: {Referring} to {Objects} in {Photographs} of {Natural} {Scenes}}.
\newblock In \emph{Proceedings of the 2014 {Conference} on {Empirical} {Methods} in {Natural} {Language} {Processing}, {EMNLP} 2014, {October} 25-29, 2014, {Doha}, {Qatar}, {A} meeting of {SIGDAT}, a {Special} {Interest} {Group} of the {ACL}}, pages 787--798. ACL.

\bibitem[{Keskar et~al.(2019)Keskar, McCann, Varshney, Xiong, and Socher}]{keskar_ctrl_2019}
Nitish~Shirish Keskar, Bryan McCann, Lav~R. Varshney, Caiming Xiong, and Richard Socher. 2019.
\newblock \href {http://arxiv.org/abs/1909.05858} {{CTRL}: {A} {Conditional} {Transformer} {Language} {Model} for {Controllable} {Generation}}.
\newblock \emph{CoRR}, abs/1909.05858.
\newblock ArXiv: 1909.05858.

\bibitem[{Krishna et~al.(2017)Krishna, Zhu, Groth, Johnson, Hata, Kravitz, Chen, Kalantidis, Li, Shamma, Bernstein, and Fei-Fei}]{krishna_visual_2017}
Ranjay Krishna, Yuke Zhu, Oliver Groth, Justin Johnson, Kenji Hata, Joshua Kravitz, Stephanie Chen, Yannis Kalantidis, Li-Jia Li, David~A. Shamma, Michael~S. Bernstein, and Li~Fei-Fei. 2017.
\newblock \href {https://doi.org/10.1007/s11263-016-0981-7} {Visual {Genome}: {Connecting} {Language} and {Vision} {Using} {Crowdsourced} {Dense} {Image} {Annotations}}.
\newblock \emph{Int. J. Comput. Vision}, 123(1):32--73.
\newblock Place: USA Publisher: Kluwer Academic Publishers.

\bibitem[{Kuznetsova et~al.(2020)Kuznetsova, Rom, Alldrin, Uijlings, Krasin, Pont-Tuset, Kamali, Popov, Malloci, Kolesnikov, Duerig, and Ferrari}]{kuznetsova_open_2020-1}
Alina Kuznetsova, Hassan Rom, Neil Alldrin, Jasper R.~R. Uijlings, Ivan Krasin, Jordi Pont-Tuset, Shahab Kamali, Stefan Popov, Matteo Malloci, Alexander Kolesnikov, Tom Duerig, and Vittorio Ferrari. 2020.
\newblock \href {https://doi.org/10.1007/S11263-020-01316-Z} {The {Open} {Images} {Dataset} {V4}}.
\newblock \emph{Int. J. Comput. Vis.}, 128(7):1956--1981.

\bibitem[{Kwon et~al.(2023)Kwon, Li, Zhuang, Sheng, Zheng, Yu, Gonzalez, Zhang, and Stoica}]{kwon_efficient_2023}
Woosuk Kwon, Zhuohan Li, Siyuan Zhuang, Ying Sheng, Lianmin Zheng, Cody~Hao Yu, Joseph Gonzalez, Hao Zhang, and Ion Stoica. 2023.
\newblock \href {https://doi.org/10.1145/3600006.3613165} {Efficient {Memory} {Management} for {Large} {Language} {Model} {Serving} with {PagedAttention}}.
\newblock In \emph{Proceedings of the 29th {Symposium} on {Operating} {Systems} {Principles}, {SOSP} 2023, {Koblenz}, {Germany}, {October} 23-26, 2023}, pages 611--626. ACM.

\bibitem[{Laurençon et~al.(2023)Laurençon, Saulnier, Tronchon, Bekman, Singh, Lozhkov, Wang, Karamcheti, Rush, Kiela, Cord, and Sanh}]{laurencon_obelisc_2023}
Hugo Laurençon, Lucile Saulnier, Léo Tronchon, Stas Bekman, Amanpreet Singh, Anton Lozhkov, Thomas Wang, Siddharth Karamcheti, Alexander~M. Rush, Douwe Kiela, Matthieu Cord, and Victor Sanh. 2023.
\newblock \href {https://doi.org/10.48550/ARXIV.2306.16527} {{OBELISC}: {An} {Open} {Web}-{Scale} {Filtered} {Dataset} of {Interleaved} {Image}-{Text} {Documents}}.
\newblock \emph{CoRR}, abs/2306.16527.
\newblock ArXiv: 2306.16527.

\bibitem[{Leng et~al.(2023)Leng, Zhang, Chen, Li, Lu, Miao, and Bing}]{leng_mitigating_2023}
Sicong Leng, Hang Zhang, Guanzheng Chen, Xin Li, Shijian Lu, Chunyan Miao, and Lidong Bing. 2023.
\newblock \href {https://doi.org/10.48550/ARXIV.2311.16922} {Mitigating {Object} {Hallucinations} in {Large} {Vision}-{Language} {Models} through {Visual} {Contrastive} {Decoding}}.
\newblock \emph{CoRR}, abs/2311.16922.
\newblock ArXiv: 2311.16922.

\bibitem[{Li et~al.(2023{\natexlab{a}})Li, Li, Savarese, and Hoi}]{li_blip-2_2023}
Junnan Li, Dongxu Li, Silvio Savarese, and Steven C.~H. Hoi. 2023{\natexlab{a}}.
\newblock \href {https://doi.org/10.48550/arXiv.2301.12597} {{BLIP}-2: {Bootstrapping} {Language}-{Image} {Pre}-training with {Frozen} {Image} {Encoders} and {Large} {Language} {Models}}.
\newblock \emph{CoRR}, abs/2301.12597.
\newblock ArXiv: 2301.12597.

\bibitem[{Li et~al.(2023{\natexlab{b}})Li, Du, Zhou, Wang, Zhao, and Wen}]{li_evaluating_2023}
Yifan Li, Yifan Du, Kun Zhou, Jinpeng Wang, Wayne~Xin Zhao, and Ji-Rong Wen. 2023{\natexlab{b}}.
\newblock \href {https://doi.org/10.48550/arXiv.2305.10355} {Evaluating {Object} {Hallucination} in {Large} {Vision}-{Language} {Models}}.
\newblock \emph{CoRR}, abs/2305.10355.
\newblock ArXiv: 2305.10355.

\bibitem[{Lin et~al.(2014)Lin, Maire, Belongie, Hays, Perona, Ramanan, Dollár, and Zitnick}]{lin_microsoft_2014}
Tsung-Yi Lin, Michael Maire, Serge~J. Belongie, James Hays, Pietro Perona, Deva Ramanan, Piotr Dollár, and C.~Lawrence Zitnick. 2014.
\newblock \href {https://doi.org/10.1007/978-3-319-10602-1_48} {Microsoft {COCO}: {Common} {Objects} in {Context}}.
\newblock In \emph{Computer {Vision} - {ECCV} 2014 - 13th {European} {Conference}, {Zurich}, {Switzerland}, {September} 6-12, 2014, {Proceedings}, {Part} {V}}, volume 8693 of \emph{Lecture {Notes} in {Computer} {Science}}, pages 740--755. Springer.

\bibitem[{Liu et~al.(2023{\natexlab{a}})Liu, Lin, Li, Wang, Yacoob, and Wang}]{liu_aligning_2023}
Fuxiao Liu, Kevin Lin, Linjie Li, Jianfeng Wang, Yaser Yacoob, and Lijuan Wang. 2023{\natexlab{a}}.
\newblock \href {https://doi.org/10.48550/ARXIV.2306.14565} {Aligning {Large} {Multi}-{Modal} {Model} with {Robust} {Instruction} {Tuning}}.
\newblock \emph{CoRR}, abs/2306.14565.
\newblock ArXiv: 2306.14565.

\bibitem[{Liu et~al.(2023{\natexlab{b}})Liu, Li, Li, and Lee}]{liu_improved_2023}
Haotian Liu, Chunyuan Li, Yuheng Li, and Yong~Jae Lee. 2023{\natexlab{b}}.
\newblock \href {https://doi.org/10.48550/ARXIV.2310.03744} {Improved {Baselines} with {Visual} {Instruction} {Tuning}}.
\newblock \emph{CoRR}, abs/2310.03744.
\newblock ArXiv: 2310.03744.

\bibitem[{Liu et~al.(2023{\natexlab{c}})Liu, Li, Wu, and Lee}]{liu_visual_2023}
Haotian Liu, Chunyuan Li, Qingyang Wu, and Yong~Jae Lee. 2023{\natexlab{c}}.
\newblock \href {https://doi.org/10.48550/arXiv.2304.08485} {Visual {Instruction} {Tuning}}.
\newblock \emph{CoRR}, abs/2304.08485.
\newblock ArXiv: 2304.08485.

\bibitem[{Loshchilov and Hutter(2019)}]{loshchilov_decoupled_2019}
Ilya Loshchilov and Frank Hutter. 2019.
\newblock \href {https://openreview.net/forum?id=Bkg6RiCqY7} {Decoupled {Weight} {Decay} {Regularization}}.
\newblock In \emph{7th {International} {Conference} on {Learning} {Representations}, {ICLR} 2019, {New} {Orleans}, {LA}, {USA}, {May} 6-9, 2019}. OpenReview.net.

\bibitem[{Lu et~al.(2023)Lu, Rao, Chen, Guo, Zhang, Sun, Yang, and Yang}]{lu_evaluation_2023}
Jiaying Lu, Jinmeng Rao, Kezhen Chen, Xiaoyuan Guo, Yawen Zhang, Baochen Sun, Carl~J. Yang, and Jie Yang. 2023.
\newblock \href {https://doi.org/10.48550/ARXIV.2309.04041} {Evaluation and {Mitigation} of {Agnosia} in {Multimodal} {Large} {Language} {Models}}.
\newblock \emph{CoRR}, abs/2309.04041.
\newblock ArXiv: 2309.04041.

\bibitem[{Mao et~al.(2016)Mao, Huang, Toshev, Camburu, Yuille, and Murphy}]{mao_generation_2016}
Junhua Mao, Jonathan Huang, Alexander Toshev, Oana Camburu, Alan~L. Yuille, and Kevin Murphy. 2016.
\newblock \href {https://doi.org/10.1109/CVPR.2016.9} {Generation and {Comprehension} of {Unambiguous} {Object} {Descriptions}}.
\newblock In \emph{2016 {IEEE} {Conference} on {Computer} {Vision} and {Pattern} {Recognition}, {CVPR} 2016, {Las} {Vegas}, {NV}, {USA}, {June} 27-30, 2016}, pages 11--20. IEEE Computer Society.

\bibitem[{{OpenAI}(2023)}]{openai_gpt-4_2023}
{OpenAI}. 2023.
\newblock \href {https://doi.org/10.48550/arXiv.2303.08774} {{GPT}-4 {Technical} {Report}}.
\newblock \emph{CoRR}, abs/2303.08774.
\newblock ArXiv: 2303.08774.

\bibitem[{Peng et~al.(2023)Peng, Wang, Dong, Hao, Huang, Ma, and Wei}]{peng_kosmos-2_2023}
Zhiliang Peng, Wenhui Wang, Li~Dong, Yaru Hao, Shaohan Huang, Shuming Ma, and Furu Wei. 2023.
\newblock \href {https://doi.org/10.48550/ARXIV.2306.14824} {Kosmos-2: {Grounding} {Multimodal} {Large} {Language} {Models} to the {World}}.
\newblock \emph{CoRR}, abs/2306.14824.
\newblock ArXiv: 2306.14824.

\bibitem[{Plummer et~al.(2015)Plummer, Wang, Cervantes, Caicedo, Hockenmaier, and Lazebnik}]{plummer_flickr30k_2015}
Bryan~A. Plummer, Liwei Wang, Chris~M. Cervantes, Juan~C. Caicedo, Julia Hockenmaier, and Svetlana Lazebnik. 2015.
\newblock \href {https://doi.org/10.1109/ICCV.2015.303} {Flickr30k {Entities}: {Collecting} {Region}-to-{Phrase} {Correspondences} for {Richer} {Image}-to-{Sentence} {Models}}.
\newblock In \emph{2015 {IEEE} {International} {Conference} on {Computer} {Vision}, {ICCV} 2015, {Santiago}, {Chile}, {December} 7-13, 2015}, pages 2641--2649.

\bibitem[{Pramanick et~al.(2023)Pramanick, Han, Hou, Nag, Lim, Ballas, Wang, Chellappa, and Almahairi}]{pramanick_jack_2023}
Shraman Pramanick, Guangxing Han, Rui Hou, Sayan Nag, Ser-Nam Lim, Nicolas Ballas, Qifan Wang, Rama Chellappa, and Amjad Almahairi. 2023.
\newblock Jack of {All} {Tasks}, {Master} of {Many}: {Designing} {General}-purpose {Coarse}-to-{Fine} {Vision}-{Language} {Model}.
\newblock \_eprint: 2312.12423.

\bibitem[{Radford et~al.(2021{\natexlab{a}})Radford, Kim, Hallacy, Ramesh, Goh, Agarwal, Sastry, Askell, Mishkin, Clark, Krueger, and Sutskever}]{radford_learning_2021-1}
Alec Radford, Jong~Wook Kim, Chris Hallacy, Aditya Ramesh, Gabriel Goh, Sandhini Agarwal, Girish Sastry, Amanda Askell, Pamela Mishkin, Jack Clark, Gretchen Krueger, and Ilya Sutskever. 2021{\natexlab{a}}.
\newblock \href {http://proceedings.mlr.press/v139/radford21a.html} {Learning {Transferable} {Visual} {Models} {From} {Natural} {Language} {Supervision}}.
\newblock In \emph{Proceedings of the 38th {International} {Conference} on {Machine} {Learning}, {ICML} 2021, 18-24 {July} 2021, {Virtual} {Event}}, volume 139 of \emph{Proceedings of {Machine} {Learning} {Research}}, pages 8748--8763. PMLR.

\bibitem[{Radford et~al.(2021{\natexlab{b}})Radford, Kim, Hallacy, Ramesh, Goh, Agarwal, Sastry, Askell, Mishkin, Clark, Krueger, and Sutskever}]{radford_learning_2021}
Alec Radford, Jong~Wook Kim, Chris Hallacy, Aditya Ramesh, Gabriel Goh, Sandhini Agarwal, Girish Sastry, Amanda Askell, Pamela Mishkin, Jack Clark, Gretchen Krueger, and Ilya Sutskever. 2021{\natexlab{b}}.
\newblock \href {https://arxiv.org/abs/2103.00020} {Learning {Transferable} {Visual} {Models} {From} {Natural} {Language} {Supervision}}.
\newblock \emph{arXiv preprint}, abs/2103.00020.
\newblock \_eprint: 2103.00020.

\bibitem[{Reimers and Gurevych(2019)}]{reimers_sentence-bert_2019}
Nils Reimers and Iryna Gurevych. 2019.
\newblock \href {https://doi.org/10.18653/v1/D19-1410} {Sentence-{BERT}: {Sentence} {Embeddings} using {Siamese} {BERT}-{Networks}}.
\newblock In \emph{Proceedings of the 2019 {Conference} on {Empirical} {Methods} in {Natural} {Language} {Processing} and the 9th {International} {Joint} {Conference} on {Natural} {Language} {Processing}, {EMNLP}-{IJCNLP} 2019, {Hong} {Kong}, {China}, {November} 3-7, 2019}, pages 3980--3990. Association for Computational Linguistics.

\bibitem[{Rohrbach et~al.(2018)Rohrbach, Hendricks, Burns, Darrell, and Saenko}]{rohrbach_object_2018}
Anna Rohrbach, Lisa~Anne Hendricks, Kaylee Burns, Trevor Darrell, and Kate Saenko. 2018.
\newblock \href {https://doi.org/10.18653/v1/d18-1437} {Object {Hallucination} in {Image} {Captioning}}.
\newblock In \emph{Proceedings of the 2018 {Conference} on {Empirical} {Methods} in {Natural} {Language} {Processing}, {Brussels}, {Belgium}, {October} 31 - {November} 4, 2018}, pages 4035--4045. Association for Computational Linguistics.

\bibitem[{Shao et~al.(2019)Shao, Li, Zhang, Peng, Yu, Zhang, Li, and Sun}]{shao_objects365_2019-1}
Shuai Shao, Zeming Li, Tianyuan Zhang, Chao Peng, Gang Yu, Xiangyu Zhang, Jing Li, and Jian Sun. 2019.
\newblock \href {https://doi.org/10.1109/ICCV.2019.00852} {Objects365: {A} {Large}-{Scale}, {High}-{Quality} {Dataset} for {Object} {Detection}}.
\newblock In \emph{2019 {IEEE}/{CVF} {International} {Conference} on {Computer} {Vision}, {ICCV} 2019, {Seoul}, {Korea} ({South}), {October} 27 - {November} 2, 2019}, pages 8429--8438. IEEE.

\bibitem[{Sun et~al.(2023)Sun, Shen, Cao, Liu, Li, Shen, Gan, Gui, Wang, Yang, Keutzer, and Darrell}]{sun_aligning_2023}
Zhiqing Sun, Sheng Shen, Shengcao Cao, Haotian Liu, Chunyuan Li, Yikang Shen, Chuang Gan, Liang-Yan Gui, Yu-Xiong Wang, Yiming Yang, Kurt Keutzer, and Trevor Darrell. 2023.
\newblock \href {https://doi.org/10.48550/ARXIV.2309.14525} {Aligning {Large} {Multimodal} {Models} with {Factually} {Augmented} {RLHF}}.
\newblock \emph{CoRR}, abs/2309.14525.
\newblock ArXiv: 2309.14525.

\bibitem[{Touvron et~al.(2023)Touvron, Lavril, Izacard, Martinet, Lachaux, Lacroix, Rozière, Goyal, Hambro, Azhar, Rodriguez, Joulin, Grave, and Lample}]{touvron_llama_2023}
Hugo Touvron, Thibaut Lavril, Gautier Izacard, Xavier Martinet, Marie-Anne Lachaux, Timothée Lacroix, Baptiste Rozière, Naman Goyal, Eric Hambro, Faisal Azhar, Aurélien Rodriguez, Armand Joulin, Edouard Grave, and Guillaume Lample. 2023.
\newblock \href {https://doi.org/10.48550/arXiv.2302.13971} {{LLaMA}: {Open} and {Efficient} {Foundation} {Language} {Models}}.
\newblock \emph{CoRR}, abs/2302.13971.
\newblock ArXiv: 2302.13971.

\bibitem[{Vedantam et~al.(2015)Vedantam, Zitnick, and Parikh}]{vedantam_cider_2015}
Ramakrishna Vedantam, C.~Lawrence Zitnick, and Devi Parikh. 2015.
\newblock \href {https://doi.org/10.1109/CVPR.2015.7299087} {{CIDEr}: {Consensus}-based image description evaluation}.
\newblock In \emph{{IEEE} {Conference} on {Computer} {Vision} and {Pattern} {Recognition}, {CVPR} 2015, {Boston}, {MA}, {USA}, {June} 7-12, 2015}, pages 4566--4575. IEEE Computer Society.

\bibitem[{Wang et~al.(2023{\natexlab{a}})Wang, Zhou, Xu, Shi, Zhao, Xu, Ye, Yan, Zhang, Zhu, Sang, and Tang}]{wang_evaluation_2023}
Junyang Wang, Yiyang Zhou, Guohai Xu, Pengcheng Shi, Chenlin Zhao, Haiyang Xu, Qinghao Ye, Ming Yan, Ji~Zhang, Jihua Zhu, Jitao Sang, and Haoyu Tang. 2023{\natexlab{a}}.
\newblock \href {https://doi.org/10.48550/ARXIV.2308.15126} {Evaluation and {Analysis} of {Hallucination} in {Large} {Vision}-{Language} {Models}}.
\newblock \emph{CoRR}, abs/2308.15126.
\newblock ArXiv: 2308.15126.

\bibitem[{Wang et~al.(2022)Wang, Yang, Men, Lin, Bai, Li, Ma, Zhou, Zhou, and Yang}]{wang_ofa_2022}
Peng Wang, An~Yang, Rui Men, Junyang Lin, Shuai Bai, Zhikang Li, Jianxin Ma, Chang Zhou, Jingren Zhou, and Hongxia Yang. 2022.
\newblock \href {https://proceedings.mlr.press/v162/wang22al.html} {{OFA}: {Unifying} {Architectures}, {Tasks}, and {Modalities} {Through} a {Simple} {Sequence}-to-{Sequence} {Learning} {Framework}}.
\newblock In \emph{International {Conference} on {Machine} {Learning}, {ICML} 2022, 17-23 {July} 2022, {Baltimore}, {Maryland}, {USA}}, volume 162 of \emph{Proceedings of {Machine} {Learning} {Research}}, pages 23318--23340. PMLR.

\bibitem[{Wang et~al.(2023{\natexlab{b}})Wang, Lv, Yu, Hong, Qi, Wang, Ji, Yang, Zhao, Song, Xu, Xu, Li, Dong, Ding, and Tang}]{wang_cogvlm_2023}
Weihan Wang, Qingsong Lv, Wenmeng Yu, Wenyi Hong, Ji~Qi, Yan Wang, Junhui Ji, Zhuoyi Yang, Lei Zhao, Xixuan Song, Jiazheng Xu, Bin Xu, Juanzi Li, Yuxiao Dong, Ming Ding, and Jie Tang. 2023{\natexlab{b}}.
\newblock \href {https://doi.org/10.48550/ARXIV.2311.03079} {{CogVLM}: {Visual} {Expert} for {Pretrained} {Language} {Models}}.
\newblock \emph{CoRR}, abs/2311.03079.
\newblock ArXiv: 2311.03079.

\bibitem[{Xiao et~al.(2023)Xiao, Liu, Zhang, and Muennighof}]{xiao_c-pack_2023}
Shitao Xiao, Zheng Liu, Peitian Zhang, and Niklas Muennighof. 2023.
\newblock \href {https://doi.org/10.48550/ARXIV.2309.07597} {C-{Pack}: {Packaged} {Resources} {To} {Advance} {General} {Chinese} {Embedding}}.
\newblock \emph{CoRR}, abs/2309.07597.
\newblock ArXiv: 2309.07597.

\bibitem[{Yin et~al.(2023)Yin, Fu, Zhao, Xu, Wang, Sui, Shen, Li, Sun, and Chen}]{yin_woodpecker_2023}
Shukang Yin, Chaoyou Fu, Sirui Zhao, Tong Xu, Hao Wang, Dianbo Sui, Yunhang Shen, Ke~Li, Xing Sun, and Enhong Chen. 2023.
\newblock \href {https://doi.org/10.48550/ARXIV.2310.16045} {Woodpecker: {Hallucination} {Correction} for {Multimodal} {Large} {Language} {Models}}.
\newblock \emph{CoRR}, abs/2310.16045.
\newblock ArXiv: 2310.16045.

\bibitem[{You et~al.(2023)You, Zhang, Gan, Du, Zhang, Wang, Cao, Chang, and Yang}]{you_ferret_2023}
Haoxuan You, Haotian Zhang, Zhe Gan, Xianzhi Du, Bowen Zhang, Zirui Wang, Liangliang Cao, Shih-Fu Chang, and Yinfei Yang. 2023.
\newblock \href {https://doi.org/10.48550/ARXIV.2310.07704} {Ferret: {Refer} and {Ground} {Anything} {Anywhere} at {Any} {Granularity}}.
\newblock \emph{CoRR}, abs/2310.07704.
\newblock ArXiv: 2310.07704.

\bibitem[{Yu et~al.(2023)Yu, Yao, Zhang, He, Han, Cui, Hu, Liu, Zheng, Sun, and Chua}]{yu_rlhf-v_2023}
Tianyu Yu, Yuan Yao, Haoye Zhang, Taiwen He, Yifeng Han, Ganqu Cui, Jinyi Hu, Zhiyuan Liu, Hai-Tao Zheng, Maosong Sun, and Tat-Seng Chua. 2023.
\newblock \href {https://doi.org/10.48550/ARXIV.2312.00849} {{RLHF}-{V}: {Towards} {Trustworthy} {MLLMs} via {Behavior} {Alignment} from {Fine}-grained {Correctional} {Human} {Feedback}}.
\newblock \emph{CoRR}, abs/2312.00849.
\newblock ArXiv: 2312.00849.

\bibitem[{Zhai et~al.(2023{\natexlab{a}})Zhai, Yang, Zhao, Xu, Shen, Zhao, Keutzer, Li, Yan, and Fan}]{zhai_halle-switch_2023}
Bohan Zhai, Shijia Yang, Xiangchen Zhao, Chenfeng Xu, Sheng Shen, Dongdi Zhao, Kurt Keutzer, Manling Li, Tan Yan, and Xiangjun Fan. 2023{\natexlab{a}}.
\newblock \href {https://doi.org/10.48550/ARXIV.2310.01779} {{HallE}-{Switch}: {Rethinking} and {Controlling} {Object} {Existence} {Hallucinations} in {Large} {Vision} {Language} {Models} for {Detailed} {Caption}}.
\newblock \emph{CoRR}, abs/2310.01779.
\newblock ArXiv: 2310.01779.

\bibitem[{Zhai et~al.(2023{\natexlab{b}})Zhai, Mustafa, Kolesnikov, and Beyer}]{zhai_sigmoid_2023-1}
Xiaohua Zhai, Basil Mustafa, Alexander Kolesnikov, and Lucas Beyer. 2023{\natexlab{b}}.
\newblock \href {https://doi.org/10.48550/ARXIV.2303.15343} {Sigmoid {Loss} for {Language} {Image} {Pre}-{Training}}.
\newblock \emph{CoRR}, abs/2303.15343.
\newblock ArXiv: 2303.15343.

\bibitem[{Zhang et~al.(2023{\natexlab{a}})Zhang, Sun, Chen, Xiao, Shao, Zhang, Chen, and Luo}]{zhang_gpt4roi_2023}
Shilong Zhang, Peize Sun, Shoufa Chen, Min Xiao, Wenqi Shao, Wenwei Zhang, Kai Chen, and Ping Luo. 2023{\natexlab{a}}.
\newblock \href {https://doi.org/10.48550/ARXIV.2307.03601} {{GPT4RoI}: {Instruction} {Tuning} {Large} {Language} {Model} on {Region}-of-{Interest}}.
\newblock \emph{CoRR}, abs/2307.03601.
\newblock ArXiv: 2307.03601.

\bibitem[{Zhang et~al.(2023{\natexlab{b}})Zhang, Li, Cui, Cai, Liu, Fu, Huang, Zhao, Zhang, Chen, Wang, Luu, Bi, Shi, and Shi}]{zhang_sirens_2023}
Yue Zhang, Yafu Li, Leyang Cui, Deng Cai, Lemao Liu, Tingchen Fu, Xinting Huang, Enbo Zhao, Yu~Zhang, Yulong Chen, Longyue Wang, Anh~Tuan Luu, Wei Bi, Freda Shi, and Shuming Shi. 2023{\natexlab{b}}.
\newblock \href {https://doi.org/10.48550/ARXIV.2309.01219} {Siren's {Song} in the {AI} {Ocean}: {A} {Survey} on {Hallucination} in {Large} {Language} {Models}}.
\newblock \emph{CoRR}, abs/2309.01219.
\newblock ArXiv: 2309.01219.

\bibitem[{Zhao et~al.(2023{\natexlab{a}})Zhao, Lin, Zhou, Huang, Feng, and Kang}]{zhao_bubogpt_2023}
Yang Zhao, Zhijie Lin, Daquan Zhou, Zilong Huang, Jiashi Feng, and Bingyi Kang. 2023{\natexlab{a}}.
\newblock \href {https://doi.org/10.48550/ARXIV.2307.08581} {Bubogpt: Enabling visual grounding in multi-modal llms}.
\newblock \emph{CoRR}, abs/2307.08581.

\bibitem[{Zhao et~al.(2023{\natexlab{b}})Zhao, Wang, Ouyang, Dong, Wang, and He}]{zhao_beyond_2023}
Zhiyuan Zhao, Bin Wang, Linke Ouyang, Xiaoyi Dong, Jiaqi Wang, and Conghui He. 2023{\natexlab{b}}.
\newblock \href {https://doi.org/10.48550/ARXIV.2311.16839} {Beyond {Hallucinations}: {Enhancing} {LVLMs} through {Hallucination}-{Aware} {Direct} {Preference} {Optimization}}.
\newblock \emph{CoRR}, abs/2311.16839.
\newblock ArXiv: 2311.16839.

\bibitem[{Zhou et~al.(2023)Zhou, Cui, Yoon, Zhang, Deng, Finn, Bansal, and Yao}]{zhou_analyzing_2023}
Yiyang Zhou, Chenhang Cui, Jaehong Yoon, Linjun Zhang, Zhun Deng, Chelsea Finn, Mohit Bansal, and Huaxiu Yao. 2023.
\newblock \href {https://doi.org/10.48550/ARXIV.2310.00754} {Analyzing and {Mitigating} {Object} {Hallucination} in {Large} {Vision}-{Language} {Models}}.
\newblock \emph{CoRR}, abs/2310.00754.
\newblock ArXiv: 2310.00754.

\bibitem[{Zhu et~al.(2023)Zhu, Chen, Shen, Li, and Elhoseiny}]{zhu_minigpt-4_2023}
Deyao Zhu, Jun Chen, Xiaoqian Shen, Xiang Li, and Mohamed Elhoseiny. 2023.
\newblock \href {https://doi.org/10.48550/arXiv.2304.10592} {{MiniGPT}-4: {Enhancing} {Vision}-{Language} {Understanding} with {Advanced} {Large} {Language} {Models}}.
\newblock \emph{CoRR}, abs/2304.10592.
\newblock ArXiv: 2304.10592.

\bibitem[{Zhu et~al.(2016)Zhu, Groth, Bernstein, and Fei-Fei}]{zhu_visual7w_2016}
Yuke Zhu, Oliver Groth, Michael~S. Bernstein, and Li~Fei-Fei. 2016.
\newblock \href {https://doi.org/10.1109/CVPR.2016.540} {{Visual7W}: {Grounded} {Question} {Answering} in {Images}}.
\newblock In \emph{2016 {IEEE} {Conference} on {Computer} {Vision} and {Pattern} {Recognition}, {CVPR} 2016, {Las} {Vegas}, {NV}, {USA}, {June} 27-30, 2016}, pages 4995--5004. IEEE Computer Society.

\end{thebibliography}
\bibliographystyle{acl_natbib}

\appendix

\section{Training and  Details}
\label{sec:appendix:train}

All models were trained on a single NVIDIA RTX3090s card, with training duration ranging between 2-4 GPU days, depending on the training task mix. We train for one epoch (on the concatenation of corpora from all tasks, as all tasks are---from the low-level technical point of view---instances of causal language modeling, i.e., next token prediction) with AdamW optimizer \cite{loshchilov_decoupled_2019} and a cosine schedule. For LoRA, we set $r=64, \alpha=128$.
During pre-training, where only the parameters of the alignment module are updated, we use
batch size 32, learning rate 0.001, and weight decay 0.
For training on the task mix, we use learning rate 2e-4, weight decay 0, and batch size 16/32/64 for Vicuna/Phi-3/Llama-3 (achieved with gradient accumulation).
 
For generation (i.e., inference), we use greedy decoding with a repetition penalty \cite{keskar_ctrl_2019} of 1.15 to avoid degenerative repetitions in long caption generation. We use one fixed prompt per task (see Table~\ref{tab:prompts}) both in training and at inference (for the subset of tasks on which we evaluate). 

We encode bounding boxes with 2 significant digits (, e.g., $[0.10, 0.05, 0.64, 1.00]$).
For grounded captions where multiple bounding boxes are needed (e.g., for something like ``three zebras''), we follow \citet{plummer_flickr30k_2015} and combine the coordinates with semicolons in the same brackets (, e.g., $[0.10, 0.05, 0.64, 1.00; 0.50, 0.15, 0.64, 1.00]$).
If we would have more than three boxes in brackets, we instead create a single bounding box covering all boxes to limit the final sequence length.

\begin{table}[]
    \centering
     \def\arraystretch{0.97}
     \resizebox{0.99\linewidth}{!}{
    \begin{tabular}{l p{5cm}}
    \toprule
    \bf Task & \bf Prompt \\
    \midrule
     Standard Caption    &  Briefly describe the image. \\
     Long Caption    & Describe the image in detail. \\
     Grounded Caption    & Describe the image and include the bounding box coordinates for every mentioned object. \\
     VQA (POPE)   &  QUESTION Answer with yes or no.\\
     Referring Expression & Give the bounding box coordinates for the region described as "DESCRIPTION". \\
     Referring Generation &  Briefly describe the region [x1, y1, x2, y2]. \\
     \bottomrule
    \end{tabular}
    }
    \caption{Prompts used for training and inference.}
    \label{tab:prompts}
\end{table}

\section{Long Captions}
\label{sec:appendix:long}
\begin{table}[t]
    \centering
    \footnotesize
    
     \def\arraystretch{0.97}
     \resizebox{0.99\linewidth}{!}{
\begin{tabular}{l rrrr}
\toprule
   \bf Model &  \bf \#Words &  \bf CHAIR$_i\downarrow$ &  \bf Coverage$\uparrow$ &  \bf Objects \\
\cmidrule(lr){2-5}
    \texttt{Llama-3} \texttt{Base} &     94.46 &   30.78 &   44.45 &   7.44 \\
    \texttt{Llama-3} \texttt{+GC} &    100.61 &   31.74 &   44.80 &   8.08 \\
\texttt{Llama-3} \texttt{+RE} &    100.39 &   29.08 &   43.66 &   7.57 \\
\texttt{Llama-3} \texttt{+RE+GC} &    103.75 &   26.42 &   43.86 &   7.66 \\
\cmidrule(lr){2-5}
      \texttt{Phi-3} \texttt{Base} &     99.17 &   27.18 &   46.16 &   7.00 \\
      \texttt{Phi-3} \texttt{+GC} &     94.33 &   25.69 &   45.45 &   6.97 \\
  \texttt{Phi-3} \texttt{+RE} &     97.09 &   27.75 &   45.20 &   6.85 \\
  \texttt{Phi-3} \texttt{+RE+GC} &     96.55 &   27.74 &   45.69 &   7.12 \\
\cmidrule(lr){2-5}
\texttt{Vicuna} \texttt{Base} &     93.91 &   26.10 &   45.12 &   7.18 \\
\texttt{Vicuna} \texttt{+GC} &     89.69 &   25.61 &   44.42 &   7.25 \\
   \texttt{Vicuna} \texttt{+RE} &     96.45 &   28.76 &   43.20 &   6.94 \\
   \texttt{Vicuna} \texttt{+RE+GC} &     90.18 &   26.06 &   44.10 &   7.28 \\
\cmidrule(lr){2-5}
   \texttt{Vicuna (Perc.)} \texttt{Base} &     93.98 &   31.52 &   41.18 &   7.02 \\
   \texttt{Vicuna (Perc.)} \texttt{+GC} &     92.64 &   31.28 &   40.67 &   7.24 \\
      \texttt{Vicuna (Perc.)} \texttt{+RE} &     96.39 &   32.79 &   40.15 &   7.08 \\
      \texttt{Vicuna (Perc.)} \texttt{+RE+GC} &     96.14 &   35.10 &   41.32 &   7.94 \\

\bottomrule
\end{tabular}
}
    \caption{Results for long captions on 
    Objects365. We report the average number of words and CHAIR metrics.
    Results with FaithScore and on MSCOCO are qualitatively the same so we omit them for brevity.}
    \label{table:experiments:long}
    \vspace{-1em}
\end{table}

%
Table~\ref{table:experiments:long} shows long captioning results.
For brevity, we only report the results for Objects365 with CHAIR(-MEN): for MSCOCO and FaithScore the results are qualitatively the same.
Overall, the differences between model variants are negligible similar to the standard captions. 
The grounding objectives (\texttt{+RE} and \texttt{+GC}) thus does not seem to affect long captions. This again questions the extent to which improved fine-grained image understanding from grounding actually transfers to hallucination reduction in open generation.

\section{CHAIR and CHAIR-MEN}
\label{sec:appendix:chair}

We report results based on our CHAIR-MEN approach in the main paper. 
In the following, we compare them against vanilla CHAIR results based on the string matching method.
In Table~\ref{table:appendix:chair}, we report string-matching CHAIR results for MSCOCO, which can be compared to Table~\ref{table:experiments:standard} (standard captions), Table~\ref{table:experiments:grounded} (grounded captions), and Table~\ref{table:experiments:long} (long captions).

We find that results with CHAIR-MEN are highly proportional to CHAIR.
This validates CHAIR-MEN as an alternative approach for identifying hallucinated objects and opens up the extension to other datasets like Objects365.

\begin{table}[]
\begin{subtable}{0.99\linewidth}
     \def\arraystretch{0.97}
     \resizebox{0.99\linewidth}{!}{
\begin{tabular}{lrrr}
\toprule\
\bf  Model &  \bf CHAIR$_i\downarrow$  & \bf Coverage$\uparrow$   &  \bf Objects \\
\midrule
        \texttt{Llama-3} \texttt{Base} &     4.36 &           58.84 &           1.62 \\
    \texttt{Llama-3} \texttt{+GC} &     4.12 &           57.30 &           1.57 \\
       \texttt{Llama-3} \texttt{+RE} &     4.36 &           58.06 &           1.61 \\
\texttt{Llama-3} \texttt{+RE+GC} &     5.30 &           59.41 &           1.68 \\
             \texttt{Phi-3} \texttt{Base} &     4.26 &           60.39 &           1.70 \\
      \texttt{Phi-3} \texttt{+GC} &     4.39 &           59.79 &           1.67 \\
         \texttt{Phi-3} \texttt{+RE} &     4.41 &           59.73 &           1.69 \\
  \texttt{Phi-3} \texttt{+RE+GC} &     4.44 &           59.21 &           1.67 \\
                     \texttt{Vicuna} \texttt{Base} &     4.45 &           58.62 &           1.62 \\
              \texttt{Vicuna} \texttt{+GC} &     3.46 &           57.74 &           1.55 \\
                 \texttt{Vicuna} \texttt{+RE} &     4.14 &           57.78 &           1.59 \\
          \texttt{Vicuna} \texttt{+RE+GC} &     3.92 &           56.80 &           1.55 \\
                 \texttt{Vicuna (Perc.)} \texttt{Base} &     5.66 &           57.50 &           1.60 \\
          \texttt{Vicuna (Perc.)} \texttt{+GC} &     4.87 &           57.10 &           1.55 \\
             \texttt{Vicuna (Perc.)} \texttt{+RE} &     5.38 &           57.57 &           1.60 \\
      \texttt{Vicuna (Perc.)} \texttt{+RE+GC} &     6.08 &           58.33 &           1.62 \\
\bottomrule
\end{tabular}
}
\caption{MSCOCO Standard Captions}
\end{subtable}

\begin{subtable}{0.99\linewidth}
     \def\arraystretch{0.97}
     \resizebox{0.99\linewidth}{!}{
\begin{tabular}{lrrr}
\toprule\
\bf  Model &  \bf CHAIR$_i\downarrow$  & \bf Coverage$\uparrow$   &  \bf Objects \\
\midrule
   \texttt{Llama-3} \texttt{+GC} &     4.32 &           53.21 &           1.41 \\
\texttt{Llama-3} \texttt{+RE+GC} &     5.21 &           54.71 &           1.48 \\
      \texttt{Phi-3} \texttt{+GC} &     4.03 &           54.61 &           1.44 \\
  \texttt{Phi-3} \texttt{+RE+GC} &     3.49 &           54.28 &           1.43 \\
              \texttt{Vicuna} \texttt{+GC} &     3.98 &           52.66 &           1.38 \\
          \texttt{Vicuna} \texttt{+RE+GC} &     3.33 &           53.54 &           1.41 \\
          \texttt{Vicuna (Perc.)} \texttt{+GC} &     4.78 &           52.29 &           1.38 \\
      \texttt{Vicuna (Perc.)} \texttt{+RE+GC} &     6.65 &           52.37 &           1.41 \\
\bottomrule
\end{tabular}
}
\caption{MSCOCO Grounded Captions}
\end{subtable}

\begin{subtable}{0.99\linewidth}
     \def\arraystretch{0.97}
     \resizebox{0.99\linewidth}{!}{
\begin{tabular}{lrrr}
\toprule\
\bf  Model &  \bf CHAIR$_i\downarrow$  & \bf Coverage$\uparrow$   &  \bf Objects \\
\midrule
            \texttt{Llama-3} \texttt{Base} &    23.45 &           80.62 &           7.10 \\
    \texttt{Llama-3} \texttt{+GC} &    24.54 &           80.02 &           7.62 \\
       \texttt{Llama-3} \texttt{+RE} &    23.22 &           79.37 &           7.55 \\
\texttt{Llama-3} \texttt{+RE+GC} &    20.63 &           79.23 &           7.20 \\
             \texttt{Phi-3} \texttt{Base} &    20.92 &           81.05 &           6.28 \\
      \texttt{Phi-3} \texttt{+GC} &    18.10 &           78.89 &           6.13 \\
         \texttt{Phi-3} \texttt{+RE} &    21.01 &           79.32 &           5.82 \\
  \texttt{Phi-3} \texttt{+RE+GC} &    22.16 &           79.82 &           6.31 \\
                     \texttt{Vicuna} \texttt{Base} &    17.54 &           80.17 &           6.51 \\
              \texttt{Vicuna} \texttt{+GC} &    17.70 &           78.76 &           6.33 \\
                 \texttt{Vicuna} \texttt{+RE} &    18.27 &           79.59 &           6.16 \\
          \texttt{Vicuna} \texttt{+RE+GC} &    18.20 &           78.68 &           6.49 \\
                 \texttt{Vicuna (Perc.)} \texttt{Base} &    23.35 &           77.82 &           6.71 \\
          \texttt{Vicuna (Perc.)} \texttt{+GC} &    22.19 &           77.11 &           6.76 \\
             \texttt{Vicuna (Perc.)} \texttt{+RE} &    22.74 &           77.85 &           6.67 \\
      \texttt{Vicuna (Perc.)} \texttt{+RE+GC} &    24.83 &           78.09 &           7.31 \\
\bottomrule
\end{tabular}
}
\caption{MSCOCO Long Captions}
\end{subtable}
\caption{CHAIR results for MSCOCO using the classic string-matching approach.
}
\label{table:appendix:chair}
\end{table}

\end{document}